\documentclass{article}

\usepackage{microtype}
\usepackage{graphicx}
\usepackage{caption}
\usepackage{subcaption}
\usepackage{booktabs} 

\usepackage{adjustbox}

\usepackage{hyperref}

\newcommand{\pde}{{\bm{\Delta}}}
\newcommand{\pr}{{\mathsf{pr}}}

\usepackage{amssymb}
\let\emptyset\varnothing

\usepackage{amsmath,amsfonts,bm}

\def\eqref#1{equation~\ref{#1}}

\def\1{\bm{1}}

\def\eps{{\epsilon}}

\def\rvu{{\mathbf{i}}}

\def\rvu{{\mathbf{u}}}

\def\rvx{{\mathbf{x}}}

\def\rmX{{\mathbf{X}}}

\DeclareMathAlphabet{\mathsfit}{\encodingdefault}{\sfdefault}{m}{sl}
\SetMathAlphabet{\mathsfit}{bold}{\encodingdefault}{\sfdefault}{bx}{n}

\def\gA{{\mathcal{A}}}

\def\gD{{\mathcal{D}}}

\def\gF{{\mathcal{F}}}

\def\gL{{\mathcal{L}}}
\def\gM{{\mathcal{M}}}

\def\gS{{\mathcal{S}}}

\def\gU{{\mathcal{U}}}

\def\gX{{\mathcal{X}}}

\def\sR{{\mathbb{R}}}

\usepackage[accepted]{icml2022}

\icmltitlerunning{Lie Point Symmetry Data Augmentation for Neural PDE Solvers}

\begin{document}

\twocolumn[
\icmltitle{Lie Point Symmetry Data Augmentation for Neural PDE Solvers}



\icmlsetsymbol{equal}{*}

\begin{icmlauthorlist}
\icmlauthor{Johannes Brandstetter$^\dagger$}{uva,linz}
\icmlauthor{Max Welling}{uva}
\icmlauthor{Daniel E. Worrall$^\ddagger$}{qc}
\end{icmlauthorlist}

\icmlaffiliation{qc}{Qualcomm AI Research, an initiative of Qualcomm Technologies, Inc $^\ddagger$now at Deepmind}
\icmlaffiliation{uva}{University of Amsterdam}
\icmlaffiliation{linz}{Johannes Kepler University Linz $^\dagger$now at Microsoft Research}

\icmlcorrespondingauthor{Johannes Brandstetter}{brandstetter@ml.jku.at}

\icmlkeywords{Machine Learning, ICML}

\vskip 0.3in
]



\printAffiliationsAndNotice{}  

\begin{abstract}
Neural networks are increasingly being used to solve \emph{partial differential equations} (PDEs), replacing slower numerical solvers. However, a critical issue is that neural PDE solvers require high-quality ground truth data, which usually must come from the very solvers they are designed to replace. Thus, we are presented with a proverbial chicken-and-egg problem. In this paper, we present a method, which can partially alleviate this problem, by improving neural PDE solver sample complexity---Lie point symmetry data augmentation (LPSDA). In the context of PDEs, it turns out that we are able to quantitatively derive an exhaustive list of data transformations, based on the \emph{Lie point symmetry group} of the PDEs in question, something not possible in other application areas. We present this framework and demonstrate how it can easily be deployed to improve neural PDE solver sample complexity by an order of magnitude.
\end{abstract}

\section{Introduction} \label{sec:introduction}
Simulation of the physical world around us is an important component of many numerate disciplines. From computational fluid dynamics to molecular modeling~\citep{lelievre2016} or astronomical simulation~\citep{Courant1967}, solving the underlying dynamical equations is usually analytically intractable. The equations themselves are regularly expressed as \emph{partial differential equations} (PDEs), solutions of which require efficient and accurate \emph{solvers}. In reality, however, classical solvers are expensive and custom designed per PDE class~\citep{quarteroni2009numerical}.

In recent years, with the reascendance of deep learning, it has become popular to learn PDE solvers~\citep{GreydanusEtAl2019, Bar_Sinai_2019, SanchezEtAl2020, thuerey2021pbdl}, circumventing the lengthy and often tedious process of solver design. But we are left with a proverbial `chicken-and-egg problem'. From where do we obtain the abundant data needed to train said neural solvers? It has to be generated with a classical solver, after all.

It seems the best we can do is to generate `groundtruth' data using a high-quality, slow `teacher' and then clone its behavior with a cheaper neural `student'---a form of so-called \emph{model order reduction}~\citep{Schilders2008}. This considered, a natural question is whether we can train student models with less of this `expensive' data. More precisely, how do we improve solver \emph{sample complexity}? One potential route is to exploit symmetries of the loss with respect to input transformations, building in equivariance \citep{WangWY21}. However, the design of equivariant layers is a difficult task.

A much simpler traditional method is data augmentation~\citep{Simard2003}.  However, knowing how to augment data and to what extent is a routine question faced by ML practitioners. It turns out that PDEs are uniquely special in that it is possible to characterize the full set of permissible augmentations for each PDE. We shall define permissible augmentations as \emph{Lie point symmetries} in Section \ref{sec:background}. As such, for learnable PDE solvers, we shall show that we are able to define exactly what kinds of data augmentation we need, placing a typically intuition-based aspect of deep learning pipeline design on a firm mathematical footing. 

The contributions of this paper are:
\vspace{-1em}
\begin{itemize}
\itemsep 0em 
    \item \emph{Lie point symmetry data augmentation} (LPSDA), a mathematically-grounded treatment of data augmentation for neural PDE solvers. LPSDA comprises the full set of solution-preserving, pointwise, continuous data transformations for any (analytic) PDE.
    \item Demonstration that LPSDA improves neural PDE solver sample complexity to over an order of magnitude on a SOTA neural PDE solver, across multiple PDEs (of evolution type) and training modalities.
    \item Ablations and interpolation on the effect of individual symmetries on neural PDE solvers.
    \item Demonstration that LPSDA is faster and stabler than running a classical solver to produce the same solution trajectories and measurement of the equivariance properties of classical and neural solvers.
\end{itemize}

\section{Background} \label{sec:background}
Here we cover PDEs and their Lie point symmetries. For a textbook we suggest \citet{olver1986} as an excellent resource or \citet{oliveri2010} as a more easily digestible review paper.

\paragraph{Partial Differential Equations.}
A partial differential equation (PDE) $\pde$ specifies a relationship between a \emph{solution} $\rvu: \gX \to \sR^n$ and its derivatives at all points $\rvx$ in the domain $\gX \subset \sR^m$. It is convenient to define the \emph{prolongation} of $\rvu$ as
\begin{align}
    \pr^{(n)} \rvu = (\rvu, \rvu_\rvx, \rvu_{\rvx\rvx}, ..., \rvu_{n\rvx}) \ ,
\end{align}
where $\rvu_\rvx$ is all first-order derivatives of $\rvu$ with respect to the \emph{independent variable} $\rvx$, $\rvu_{\rvx\rvx}$ is all unique second-order derivatives of $\rvu$ (e.g.\ $\partial_{xx}\rvu, \partial_{xy}\rvu, \partial_{yx}\rvu$), and so forth up until $n$\textsuperscript{th} order. The range of the prolonged function $\pr^{(n)} \rvu$ lives in the so-called \emph{jet space} $\gU^{(n)}$. PDEs are thus represented as a system of algebra\"ic equations
\begin{align}
    \Delta (\rvx, \pr^{(n)} \left . \! \rvu \right |_\rvx ) = 0 \ , \qquad \text{for all } \rvx \in \gX \ . \label{eq:pde-algebraic}
\end{align}
By way of example, the $1+1$ dimensional Heat equation in space and time can be written 
\begin{align}
    \Delta(\rvx, \pr^{(2)}\left . \! u \right |_{\rvx}) = u_t - \alpha u_{xx}  = 0 \label{eq:heat-equation}
\end{align}
for independent variables $\rvx = (x, t)$ and thermal diffusivity $\alpha > 0$. We use unbolded type for scalar-valued variables. The set $\gS_\pde = \{(\rvx, \left . \!  \rvu^{(n)} \right |_\rvx) | \pde(\rvx, \left . \!  \rvu^{(n)} \right |_\rvx) = 0 \} \subset \gX \times \gU^{(n)}$, where $\rvu^{(n)} = \pr^{(n)} \rvu$, can be thought of as a (collection of) surface(s) in \emph{total space} $\gX \times \gU^{(n)}$, is called an \emph{algebra\"ic subvariety}~\citep[p.~79]{olver1986}. It is the graph\footnote{It is unfortunate that this nomenclature is already so overloaded in our field. Graph here refers to the surface one would draw if plotting $\rvu^{(n)}$ with pencil and paper.} of all prolonged solutions. Thinking in these geometrical terms, we are equipped to define \emph{symmetries} of PDEs. 

\paragraph{Lie Point Symmetries of PDEs.}
If a function $\rvu$ satisfies Equation~\ref{eq:pde-algebraic} at point $\rvx$, then $\rvu$ is called the \emph{solution} of $\pde$ at $\rvx$. We leave a discussion on boundary conditions for later. Symmetries are classes of transformations, which map the solution set back into itself. For solution $\rvu$ we define its image after transformation as $g\rvu$. $g$ could abstractly represent a rotation, or a shear, or a flip. The set $G$ of all solutions preserving $g$ is called the \emph{symmetry group} of $\pde$. If $g\rvu$ represents a transformed $\rvu$, then $\mathsf{pr}^{(n)}g \cdot \rvu^{(n)}$ represents the corresponding transformation of the prolongation $\rvu^{(n)}$. Then $(\rvx, \left . \!  \rvu^{(n)} \right |_\rvx) \in \gS_\pde$ implies $(g\rvx, \pr^{(n)}g \cdot \left . \!  \rvu^{(n)} \right |_\rvx) \in \gS_\pde$. Defining $g(\rvx, \left . \!  \rvu^{(n)} \right |_\rvx) = (g\rvx, \pr^{(n)} g \cdot \left . \!  \rvu^{(n)} \right |_\rvx)$ we then have
\begin{align}
    g\gS_\pde = \gS_\pde \ . \label{eq:algebraic-subvariety}
\end{align}
$g\gS_\pde$ is the set where $g$ is applied to every element of $\gS_\pde$. 

\begin{figure*}[!tb]
    \centering
    \includegraphics[width=\textwidth]{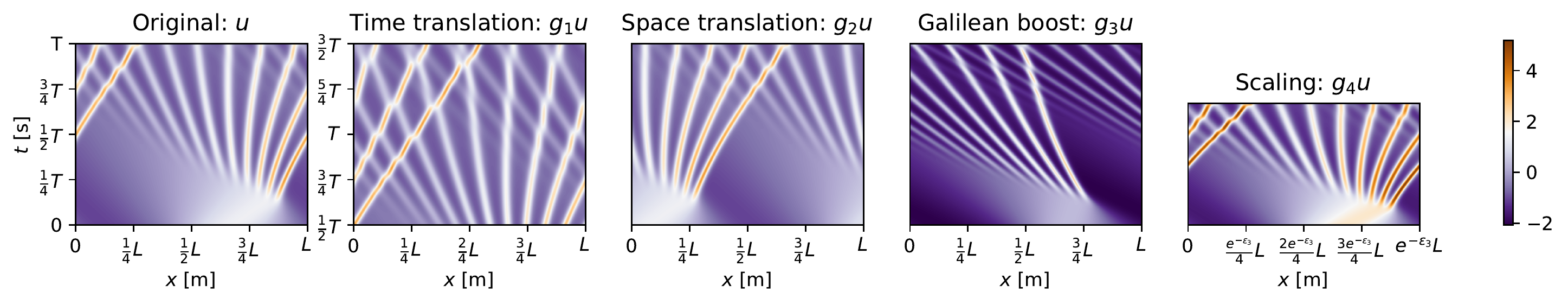}
    \caption{One-parameter Lie point symmetries of the \emph{Korteweg-de Vries equation}. The transformations are: $(g_1)$ time translation, $(g_2)$ space translation, $(g_3)$ Galilean-like boost, and $(g_4)$ scaling. The domain is periodic in the $x$-direction with units of meters and the time axis has units of seconds. Each plot has been shown for random transformations with shared spatial-scaling and colorspace.}
    \label{fig:kdv-aug}
\end{figure*}

The symmetry group of $\pde$ has the structure of a mathematical group: a set, which is closed under composition, is associative, has an identity element, and for which each element has an inverse. For PDEs, there exists a subgroup\footnote{Group within a group}, within $G$, with additional structure; namely, it is a \emph{Lie point symmetry}. Lie point symmetries are Lie groups---classes of smooth transformations with smooth inverse---that act on the $(\rvx, \left . \! \rvu \right |_{\rvx})$ representation of a function \emph{pointwise}, as
\begin{align}
    (\rvx, \left . \! \rvu \right |_{\rvx}) \overset{g}{\mapsto} (g\rvx, g \! \left . \! \rvu \right |_{\rvx})\ , \qquad \text{for all $g$ and $(\rvx, \rvu)$} \ . \label{eq:lie-point-symmetry}
\end{align}
This maps points to points, hence the name. In particular, $g$ may not depend on derivatives of $\rvu$, otherwise it is a \emph{Lie-B\"acklund transformation}, not considered in this paper. Examples of Lie point symmetries are rotations, translations, Galilean boosts, scalings, and shears. A counterexample is a blur, which neither acts pointwise, since it performs averages over neighborhoods, nor is invertible, thus not a group-structured transformation. Figure~\ref{fig:kdv-aug} shows the four Lie point symmetries of the Korteweg-de Vries equation. 

The significance of PDE symmetries has a long history dating back to Sophus Lie, who famously developed the theory of Lie groups to better solve ordinary differential equations. Most importantly for us, if we have procured a solution $\rvu$ (by any means necessary), we can generate a collection of new solutions  $\{g \rvu\}_{g\in G}$for free, given computing $g \rvu$ is relatively cheap, which we show in Section~\ref{sec:equivariance-error}.
\paragraph{One-parameter Lie groups.} 
The Lie point symmetries of a given PDE can be multidimensional and it is thus useful to break this down into modular components. Any $d$ dimensional simply-connected Lie group can be decomposed \citep[Equation~1.40]{olver1986} into a series of \emph{one-parameter transformations} $g_i: \sR \to G$ for $i=1, ..., d$, with the property $g_i(\delta)g_j(\eps) = g_i(\delta + \eps)$ if $i=j$, such that
\begin{align}
    g = g_1(\eps_1) g_2(\eps_2) \cdots g_d(\eps_d) \ . \label{eq:group-decomposition}
\end{align}
Note that $g_i(\delta)g_j(\eps) \neq g_j(\eps)g_i(\delta)$ for $i\neq j$. Each $g_i$ is a smooth, group-structured map from the real parameter $\eps_i$ on to the group, and is thus also referred to as a \emph{one-parameter group}. For us, each one-parameter group will correspond to a different kind of data augmentation we can apply. The \emph{caveat that $G$ is simply-connected} is important. It means we only consider transformations that can be continuously parameterized, for instance, 2D rotations but not reflections.

For each PDE $\pde$, a complete set of simply-connected Lie point symmetries can be derived in the form of a collection of one-parameter Lie groups \citep[\S~2.4]{olver1986}. There is extensive literature tabulating these symmetries for a vast array of common PDEs (e.g.,\ \citet{Aksenov1995CRCHO}) and for novel PDEs software exists to derive them symbolically (e.g.,\ \citet{baumann1998}).


\section{Lie Point Symmetry Data Augmentation} \label{sec:method}

Here we show how Lie point symmetries can be used as a data augmentation technique for a learned neural PDE solver. We elucidate our method with a worked example.

Given a PDE $\pde$, one can look up its Lie point symmetries as one-parameter transformations $\{g_1, ..., g_d\}$ using, say, \citet{Aksenov1995CRCHO} or \citet{baumann1998}. Theory tells us that list is exhaustive, under regularity conditions (Section~\ref{sec:exhaustiveness}). Each symmetry $g_i$ is a function of a single variable $\eps_i$, controlling the magnitude of the augmentation. For instance $\eps_i$ might correspond to a rotation angle or spatial shift. At training time, we sample each $\eps_i$ independently from an appropriate probability density $p(\eps_i)$ for $i=1,...,d$. Each density is chosen using standard techniques, such as cross-validation. At training time, one can then augment the training set $\gD$ by sampling a solution $\rvu \in \gD$ and augmenting it as
\begin{align}
    \rvu' = g_d(\eps_d) \cdots g_1(\eps_1) \rvu \ . \label{eq:sampling}
\end{align}
In words, we take $\rvu$, apply transformation $g_1(\eps_1)$, followed by $g_2(\eps_2)$, and so forth, where each $\eps_1, \eps_2, ...$ is sampled from its own density. How $\gD$ is generated is beyond the scope of this paper, but it may consist of known analytical solutions, simulation data, or real-world data. The latter two methods will contain noise meaning that $\rvu$ does not lie within the solution set, implying that $g\rvu$ does not either. We assume this noise is sufficiently small, so while $g\rvu$ may not lie exactly in the solution set, it is situated very close to it.

\paragraph{Worked Example: The Korteweg-de Vries Equation.}
By way of example, let's consider the \emph{Korteweg-de Vries} (KdV) equation \citep[p.~360]{Boussinesq}, a nonlinear scalar-valued PDE in space and time, made famous in the 1960s \citep{Zabusky1965} for exhibiting \emph{solitons}, solitary weakly-interacting ``wave-pulses''. Example solutions can be found in Figure~\ref{fig:kdv-aug}. The KdV equation is
\begin{align}
    \pde((x, t), u^{(3)}) = u_t + uu_x + u_{xxx} = 0 \ .
\end{align}
Notable is the nonlinear convective term $u u_x$ and the dispersive term $u_{xxx}$, which in traditional solvers require careful attention. We solve this equation on a domain $[0, L]$ for times $[0, T]$. We impose periodic boundary conditions $u(0, t) = u(L, t)$ and initial conditions $u(x, 0) = u_0(x)$ for all $x$ and $t$. Note $\gX = [0, L] \times [0, T]$. We discretize $\gX$ uniformly on a grid $\rmX=\{(x_i, t_j)\}_{i,j}$ of $N_x$ spatial points and $N_t$ timesteps. A typical neural PDE solver takes as input initial conditions $\rvu_0 = \{u_0(\rvx) | \rvx \in \rmX_0\}$ evaluated at $\rmX_0 = \{\rvx \in \rmX | t=0 \}$ and outputs predictions to match targets $\hat{\rvu} = \{\rvu(\rvx) | \rvx \in \rmX\}$. We augment this data as
\begin{align}
    (\rmX_0, \rvu_0, \rmX, \hat{\rvu}) \overset{g}{\mapsto} (g\rmX_0, g\rvu_0, g\rmX, g\hat{\rvu})
\end{align}
with $g$ sampled according to Equation~\ref{eq:sampling} from the Lie point symmetry group of the KdV equation. These are namely
\begin{alignat}{2}
    g_1(\eps) (x, t, u) &= (x, t + \eps, u) && \text{  time shift,}\\
    g_2(\eps) (x, t, u) &= (x + \eps, t, u) && \text{  space shift,} \label{eq:space}\\
    g_3(\eps) (x, t, u) &= (x + \eps t, t, u+\eps) && \text{  Galilean boost,}\\
    g_4(\eps) (x, t, u) &= (e^{\eps} x, e^{3\eps} t, e^{-2\eps} u) && \text{  scaling},
\end{alignat}
which can be readily found in \citet[p.~126]{olver1986}---note the sign flip in $g_3$ and $g_4$. Working with these transformations poses some technical design issues, discussed next.

\paragraph{Resampling.} A transformation, say Equation~\ref{eq:space}, maps the grid points $(x, t) \mapsto (gx, gt) = (x+\eps, t)$. This poses an issue for Eulerian solvers, where the grid $\rmX$ is always fixed. Augmented solutions $(gx, gt, gu)$ thus have to be resampled on $\rmX$. In this work, we use trigonometric interpolation in the $x$-direction. A deeper discussion is found in Appendix~\ref{sec:interpolation}. 
\vspace{-1em}
\begin{figure*}[!tb]
    \centering
    \includegraphics[width=\linewidth]{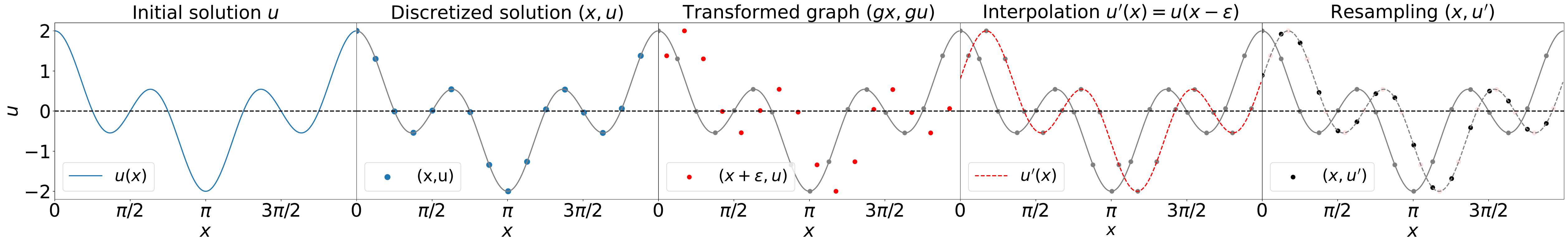}
    \caption{Schematic of the data augmentation process: From left to right we see how a solution $u$ is drawn, discretized into its graph $(x,u)$, transformed as per the symmetry (translation), interpolated with a smooth interpolant, and resampled on to the original mesh $\rmX$.}
    \label{fig:sketch}
\end{figure*}
\subsection{How complete is our symmetry list?} \label{sec:exhaustiveness}
Theory tells us our list of symmetries is exhaustive if the PDE is `nondegenerate' \citep[Theorem~2.71]{olver1986}, a technical condition that $\pde$ is of \emph{maximal rank} and \emph{locally solvable}. Intuitively, local solvability is the condition the PDE is smooth enough that points $(\rvx, \rvu^{(n)})$ satisfying $\pde$ can be interpolated locally by sufficiently smooth $\rvu$. Local solvability is implied if $\pde$ is analytic in its arguments, a corollary of the Cauchy-Kovalevskaya theorem \citep[Corollary~2.74]{olver1986}. In practice though, we need not worry ourselves with demonstrating nondegeneracy. 

\subsection{A note on boundary conditions}
The symmetries listed are defined on open neighborhoods about the origin, where $\eps$ is sufficiently small. In general, symmetries break at boundaries where open neighborhoods cannot be defined, requiring special care. We work with wrapped domains to avoid this issue. One could foresee, however, that we could sidestep the problem of boundaries by performing data augmentation on local patches, but we leave that for future work. Despite working in this restricted scenario, we are still left with sufficiently many interesting symmetries, which we demonstrate can improve generalization performance in our experiments.

\section{Experiments} \label{sec:experiments}
We ran experiments across different equations (Section~\ref{sec:pdes}), models (Section~\ref{sec:models}), training setups (Section~\ref{sec:setups}), and choices of symmetries (Section~\ref{sec:symmetry-ablation}). We tested the effectiveness of adding symmetries on raw generalization performance and survival time and demonstrate that adding LPSDA is equivalent to using larger training sets by up to a factor of $\times 16$ in size on the datasets used. We also probe the effect of LPSDA on long rollout stability (Section~\ref{sec:very-long}), and we measure the equivariance error of the learned models, comparing with classical solvers (Section~\ref{sec:equivariance-error}). 

\begin{table*}[!tb]
    \centering
    \caption{Lie point symmetries as function of $\eps \in \sR$ for studied PDEs. $\sigma: \sR \to [0,1]$ is any squashing function. Sources: KdV \citet[Ex.~2.44]{olver1986}, KS (derivative form): \citet{Cvitanovi2010OnTS}, Burgers': \citet{Hoarau2007a} and \citet[Ex.~2.42]{olver1986}.}
    \label{tab:symmetries}
    \resizebox{\linewidth}{!}{
    \begin{tabular}{c|ccccc}
        \toprule
        \textsc{Equation} & $g_1$ & $g_2$ & $g_3$ & $g_4$ & $g_\alpha$ \\
        \midrule
        KdV & $(x, t  + \eps, u)$, & $(x  + \eps, t, u)$, & $(x + \eps t, t, u+\eps)$, & $(e^{\eps} x, e^{3\eps} t, e^{-2\eps} u)$ \\
        KS & $(x, t + \eps, u)$, & $(x + \eps, t, u)$, & $(x + \eps t, t, u+\eps)$ \\
        Burgers' & $(x, t+\eps, u)$, & $(x+\eps, t, u)$, & $(x, t, u+\eps)$, & $(e^{\eps} x, e^{2\eps} t, u)$, & $ \left ( u, t, 2\nu \log \left ((1-\sigma(\epsilon))e^{\frac{1}{2\nu}u} + \sigma(\epsilon) e^{\frac{1}{2\nu}\alpha} \right ) \right )$ \\
        \bottomrule
    \end{tabular}
    }
\end{table*}
\begin{table}[!b]
    \centering
    \caption{Experimental choices for studied PDEs as per~\citet{Bar_Sinai_2019}. Ranges are sampled from uniformly.}
    \label{tab:settings}
    \resizebox{0.98\columnwidth}{!}{
    \begin{tabular}{r|cccccc}
        \toprule
        & $L$ & $T$ & $K$ & $A_k$ & $\ell_k$ & $\phi_k$ \\
        \midrule
        \textsc{KdV} & $[0.9,1.1]\cdot128$ & $[0.9,1.1]\cdot40$ & $10$ & $[-0.5,0.5]$ & $\{1,2,3\}$ & $[0,2\pi]$ \\
        \textsc{KS} & $[0.9,1.1]\cdot64$ & $[0.9,1.1]\cdot20$ & $10$ & $[-0.5,0.5]$ & $\{1,2,3\}$ & $[0,2\pi]$ \\
        \textsc{Burgers'} & $[0.9,1.1]\cdot2\pi$ & $10$ & $20$ & $[-0.5,0.5]$ & $\{3,4,5,6\}$ & $[0,2\pi]$ \\
        \bottomrule
    \end{tabular}
    }
\end{table}
\subsection{Experimental setup}
\paragraph{Equations.}
We chose to use 1D \emph{evolution equations}:
\begin{align}
    u_t - F(x, u, u_x, u_{xx}, ...) = 0 \ ,
\end{align}
as a test bed for LPSDA; although, our method can be applied to other classes of PDE. Unless otherwise stated (e.g.,\ Section~\ref{sec:pdes}), we run most ablation experiments on the KdV equation. Notwithstanding, we also consider 1) the Kuramoto-Shivashinsky (KS) equation and 2) the Burgers' equation. The KS equation is known for its chaotic behavior:
\begin{align}
    u_t + u_{xx} + u_{xxxx} + u u_x = 0 \ .
\end{align}
Above being chaotic, it is nonlinear and the biharmonic term $u_{xxxx}$ is hard to model. The Burgers' equation is 
\begin{align}
    u_t + u u_x - \nu u_{xx} = 0 \ ,
\end{align}
for viscocity $\nu \geq 0$. It exhibits shock formation, which becomes discontinuities when $\nu=0$. In our experiments, we only work with viscid systems ($\nu > 0$) where derivatives exist everywhere. Each PDE has symmetries listed in Table~\ref{tab:symmetries}. The Burgers' equation has an extra `infinite dimensional subalgebra' $g_\alpha$, which is not a Lie group symmetry, but for any two solutions $u, \alpha$, we can return a new one as $2\nu \log \left ( (1-\epsilon)e^{\frac{1}{2\nu}u(x,t)} + \epsilon e^{\frac{1}{2\nu}\alpha(x,t)} \right )$ for $\eps\in[0,1]$. Training data generation is described in Appendix~\ref{sec:data}.

\paragraph{Meshes and parameters.}
Unless otherwise stated, we set $\Omega = [0, L]$ discretized uniformly over $N_x = 256$ points with periodic boundaries and set $t \in [0, T]$ at $N_{t_\text{out}}=100$ uniformly sampled timesteps. Equation specific settings are in Table~\ref{tab:settings}. Initial conditions are sampled per \citet{Bar_Sinai_2019} from a distribution over truncated Fourier series with random coefficients $\{A_k, \ell_k, \phi_k\}_k$ as
\begin{align}
    u_0(x) = \sum_{k=1}^K A_k \sin(2 \pi \ell_k x / L + \phi_k) \ .
\end{align}

\paragraph{Neural PDE Solvers.}
We experimented with two models: a litmus test that data augmentation can benefit more than one model class. These were: a baseline 1D ResNet-like model \citep{HeEtAl2016}, and and adapted version of the Fourier Neural Operator (FNO)~\citep{li2020fourier}, which is SOTA for regular meshes with periodic boundaries. The exact architectures are in Appendix~\ref{sec:architectures}. Both models are trained in two different ways following \citet{BrandstetterEtAl2022}: (i) a \emph{neural operator method} and (ii) an \emph{autoregressive method}.

\begin{figure}[b]
    \centering
    \includegraphics[width=\linewidth]{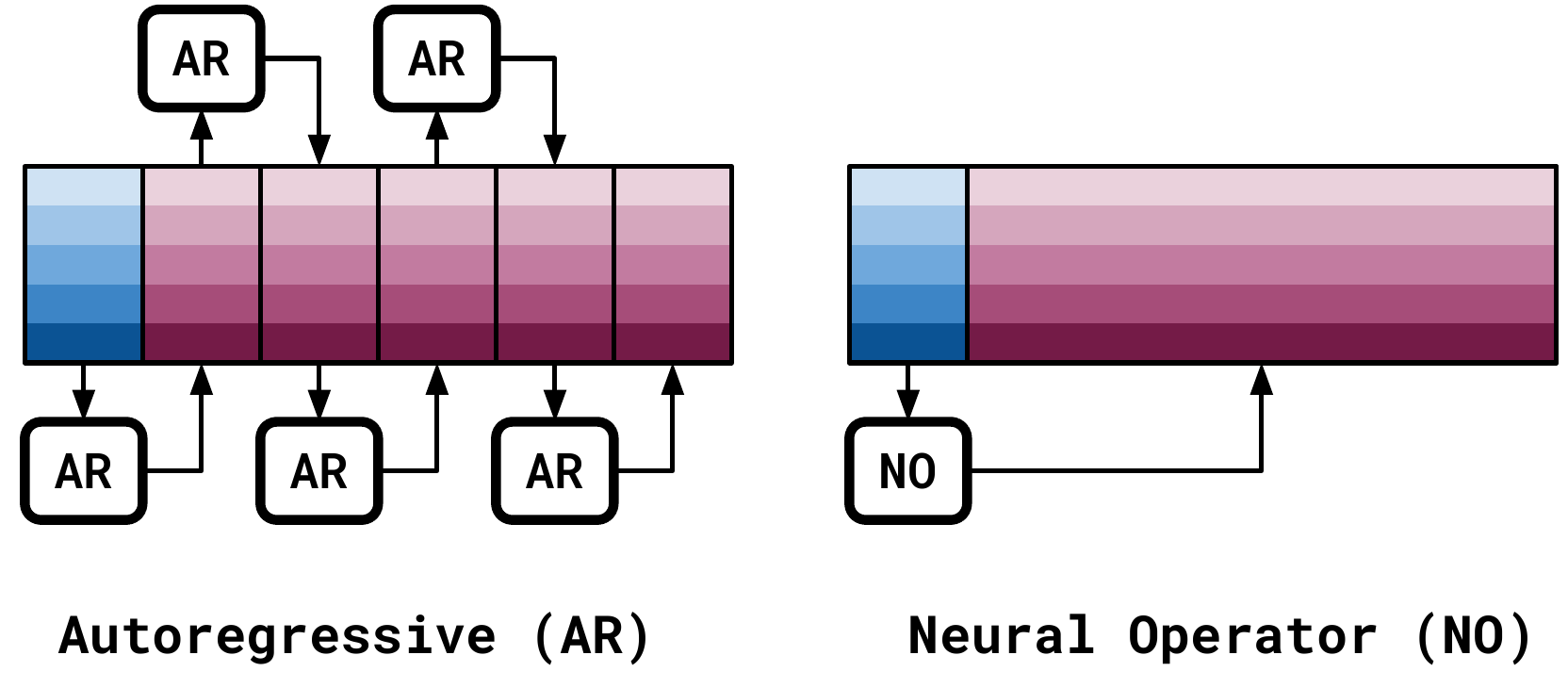}
    \caption{Autoregressive models predict the solution recurrently; whereas neural operator methods predict it directly.}
    \label{fig:schematic}
\end{figure}

\paragraph{Training methodologies.}
A neural operator (NO) is a mapping $\gM: [0, T] \times \gF \to \gF$, where $\gF$ is a (possibly infinite-dimensional) function space, trained to satisfy
\begin{align}
	\gM(t, \rvu_0) = \rvu(t) \ .
\end{align}
It maps initial conditions $\rvu_0$ directly to solutions $\rvu$ at time $t$. The flavor we use predicts all $N_{t_\text{out}}$ future times simultaneously, a technique called \emph{temporal bundling}~\citep{BrandstetterEtAl2022}. The initial conditions are also extended in time as a trajectory of the last $N_{t_\text{in}}=20$ steps. This assumes our prediction task has a fixed rollout length. An autoregressive (AR) method on the other hand solves the PDE iteratively, allowing it to run for variable length (up to multiples of the temporal bundling size). For time-dependent PDEs with time-independent coefficients, the solution is computed as
\begin{align}
	\rvu(t+\Delta t) = \gA(\Delta t, \rvu(t)) \ ,
\end{align}
where $\gA: \sR_{> 0} \times \sR^n \to \sR^n$ is the temporal update. Again, the flavor we use is a temporally bundled variant, which predicts $N_{t_\text{AR}}$ future timesteps simultaneously in each recurrent iteration. A schematic of these two training methodologies is shown in Figure~\ref{fig:schematic}. We trained in these two ways so that LPSDA is tested in different task setups. 
\begin{figure*}[htb!]
    \centering
    \begin{subfigure}[b]{0.24\textwidth}
        \centering
        \includegraphics[width=\textwidth]{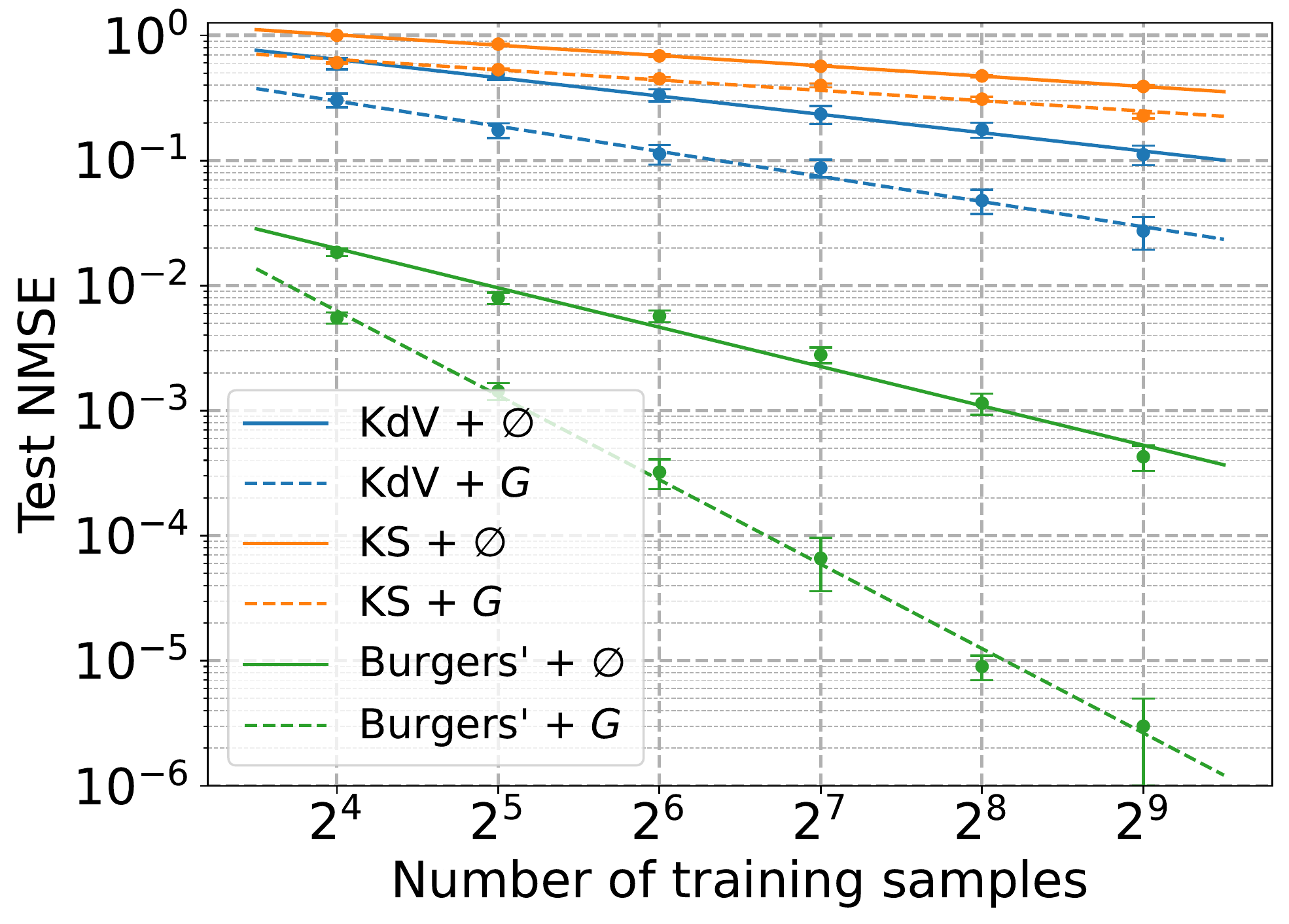}
        \caption{Different PDEs}
        \label{fig:kdv-ks}
    \end{subfigure}
    \begin{subfigure}[b]{0.24\textwidth}
        \centering
        \includegraphics[width=\textwidth]{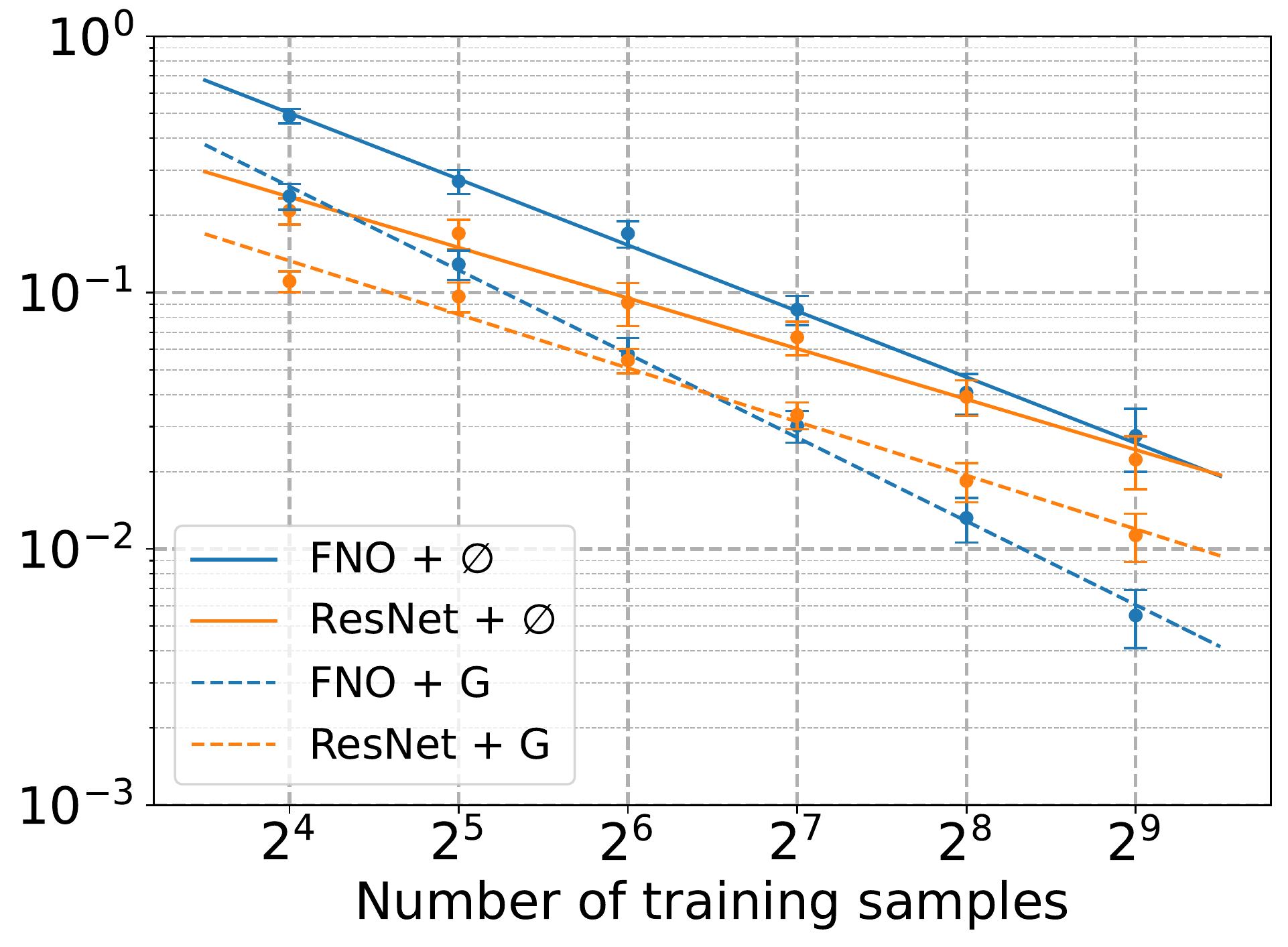}
        \caption{Different Models}
        \label{fig:kdv-model}
    \end{subfigure}
    \begin{subfigure}[b]{0.24\textwidth}
        \centering
        \includegraphics[width=\textwidth]{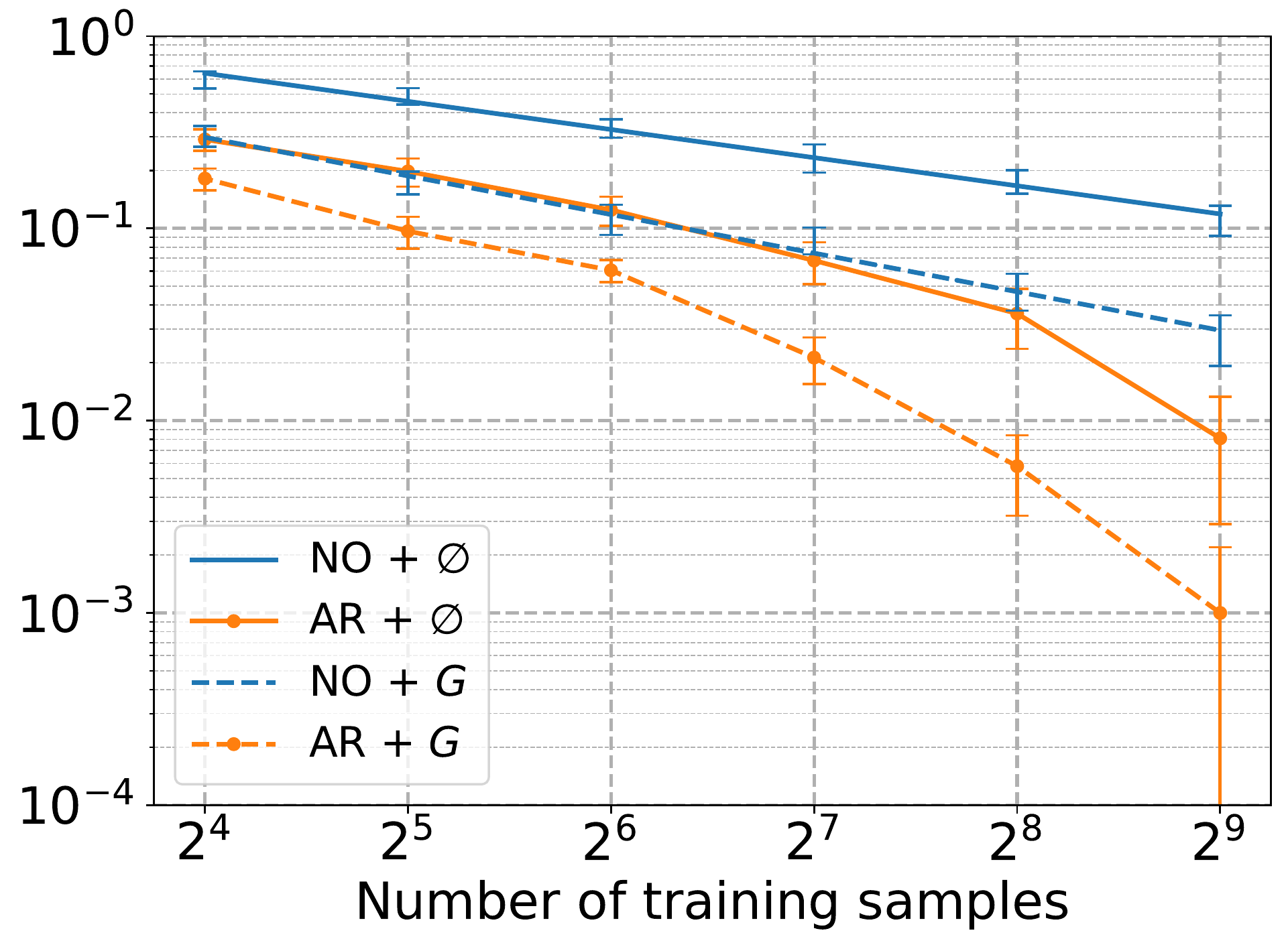}
        \caption{Training methods}
        \label{fig:kdv-method}
    \end{subfigure}
    \begin{subfigure}[b]{0.24\textwidth}
        \centering
        \includegraphics[width=\linewidth]{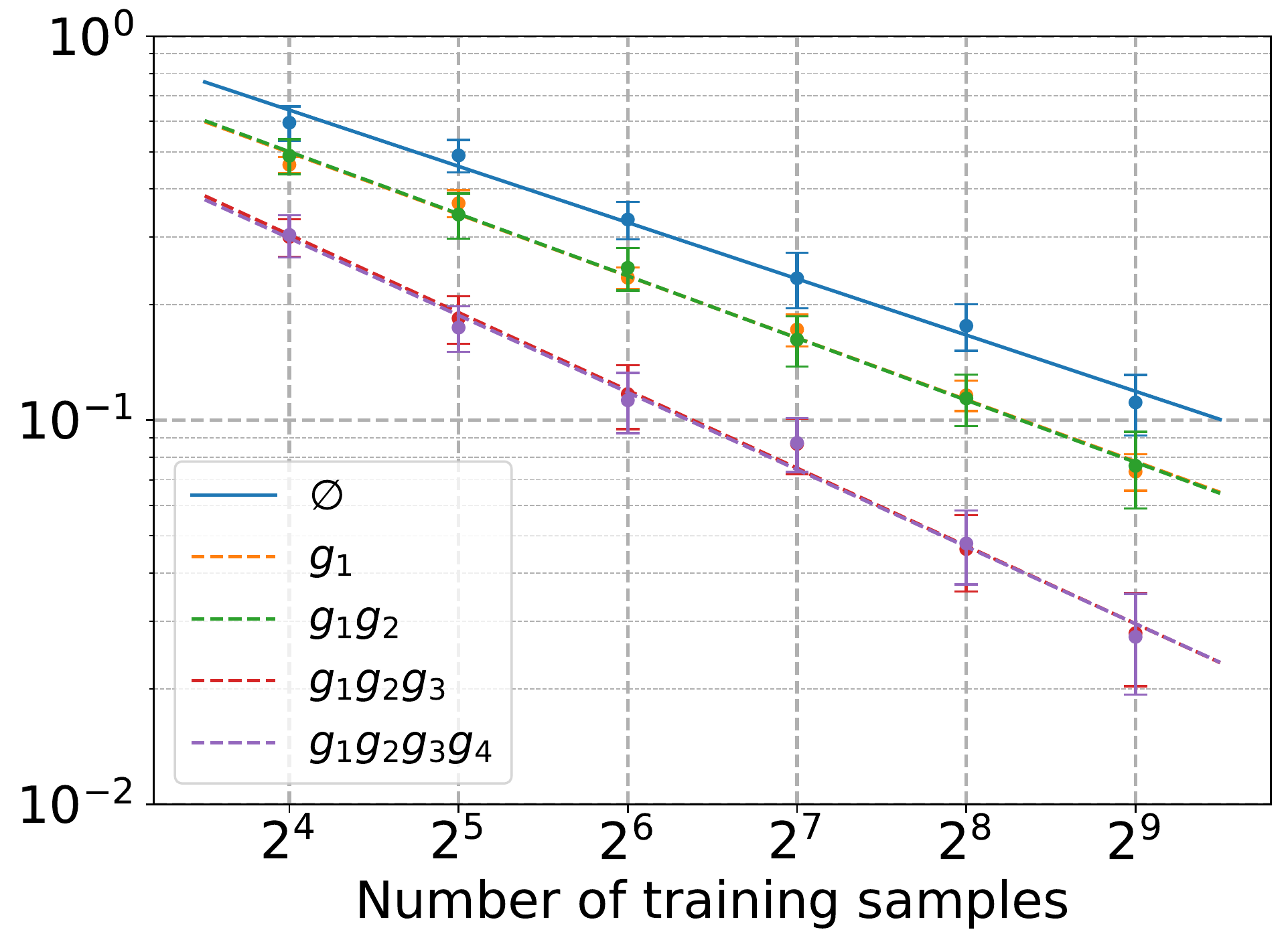}
        \caption{Combining symmetries}
        \label{fig:kdv-multiple}
    \end{subfigure}
    \caption{We compare test set normalized MSE with and without LPSDA across training set sizes. All errorbars are $\pm$ 2 std, calculated via bootstrap. A comprehensive summary of results can be found in Table \ref{tab:summary_of_results} in the appendix. (a): LPSDA improves sample complexity on the KdV and KS equations by $\times 4$ to $\times 8$, (b) LPSDA improves sample complexity on Fourier Neural Operator and Residual Networks on the KdV equation, (c) LPSDA is also effective across neural operator (NO) and autoregressive (AR) training methodologies on the KdV equation, (d) on the KdV equation time translation $(g_1)$ and Galilean boosts $(g_3)$ appear to help a lot, but space translation $(g_2)$ and scaling $(g_4)$ do not.}
\end{figure*}
\paragraph{Evaluation metrics.}
We report in terms of rollout averaged normalized MSE (NMSE), defined as
\begin{align}
    \gL = \frac{1}{N_t}\sum_{j=1}^{N_t} \frac{\Vert \rvu(t_j) - \hat{\rvu}(t_j) \Vert_2^2}{\Vert \hat{\rvu}(t_j) \Vert^2} \ ,
    \label{eq:loss}
\end{align}
where $\rvu$ is the target and $\hat{\rvu}$ is the model output. The normalized MSE is itself dimensionless and invariant to rescaling of the spatial axes. However, the average over timesteps is not invariant to time rescaling, with dimensions of error per timestep. We chose to report this error instead of error per second, because a) we can compare across different PDEs and b) we only need look at relative improvement of LPSDA on top of an arbitrary baseline without augmentation. All errors are reported with $\pm 2$ standard deviation error bars ($95 \%$ confidence intervals), computed with bootstrapping.

\begin{figure}[!b]
    \centering
    \includegraphics[width=0.325\linewidth]{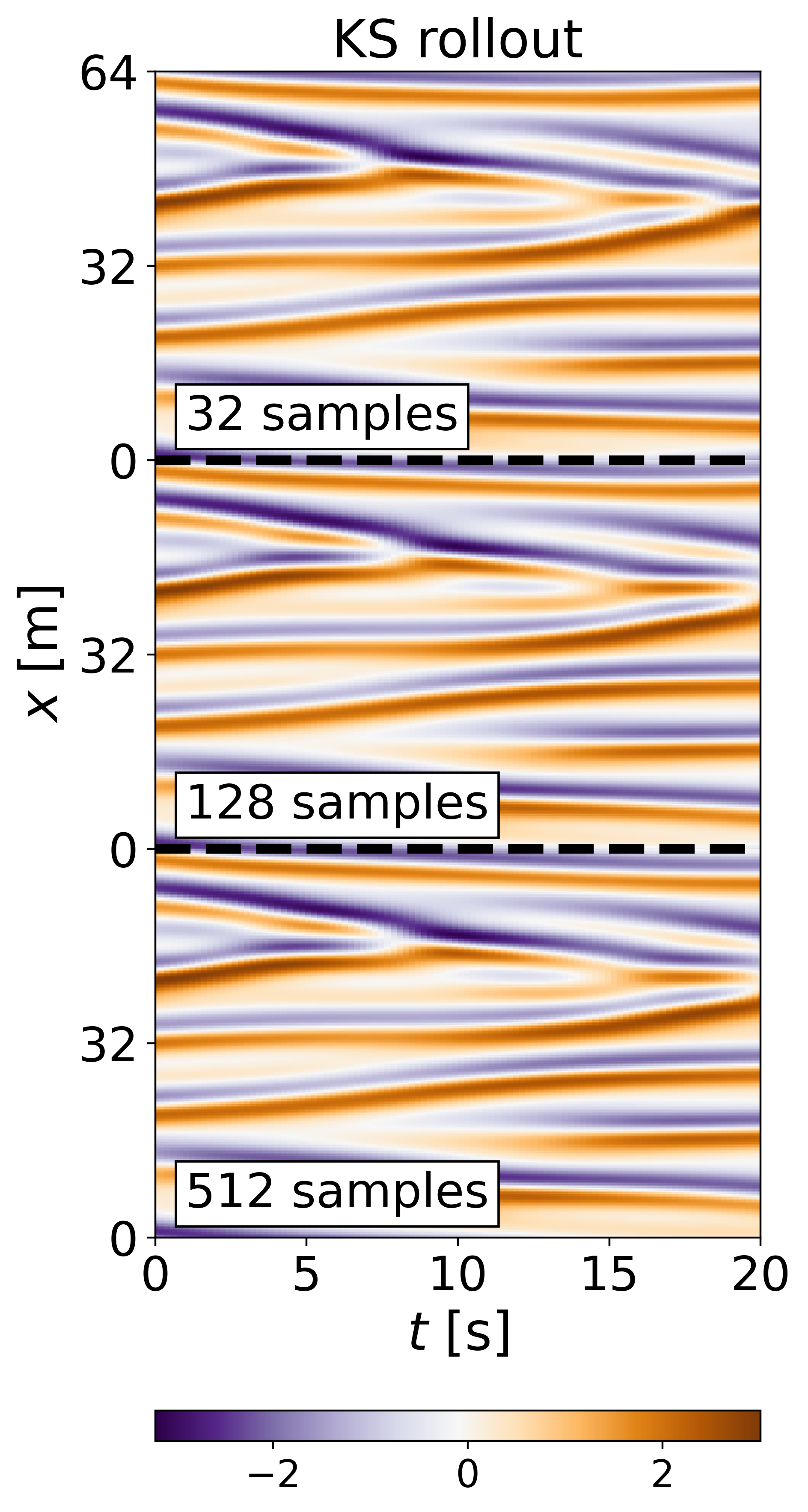}
    \includegraphics[width=0.325\linewidth]{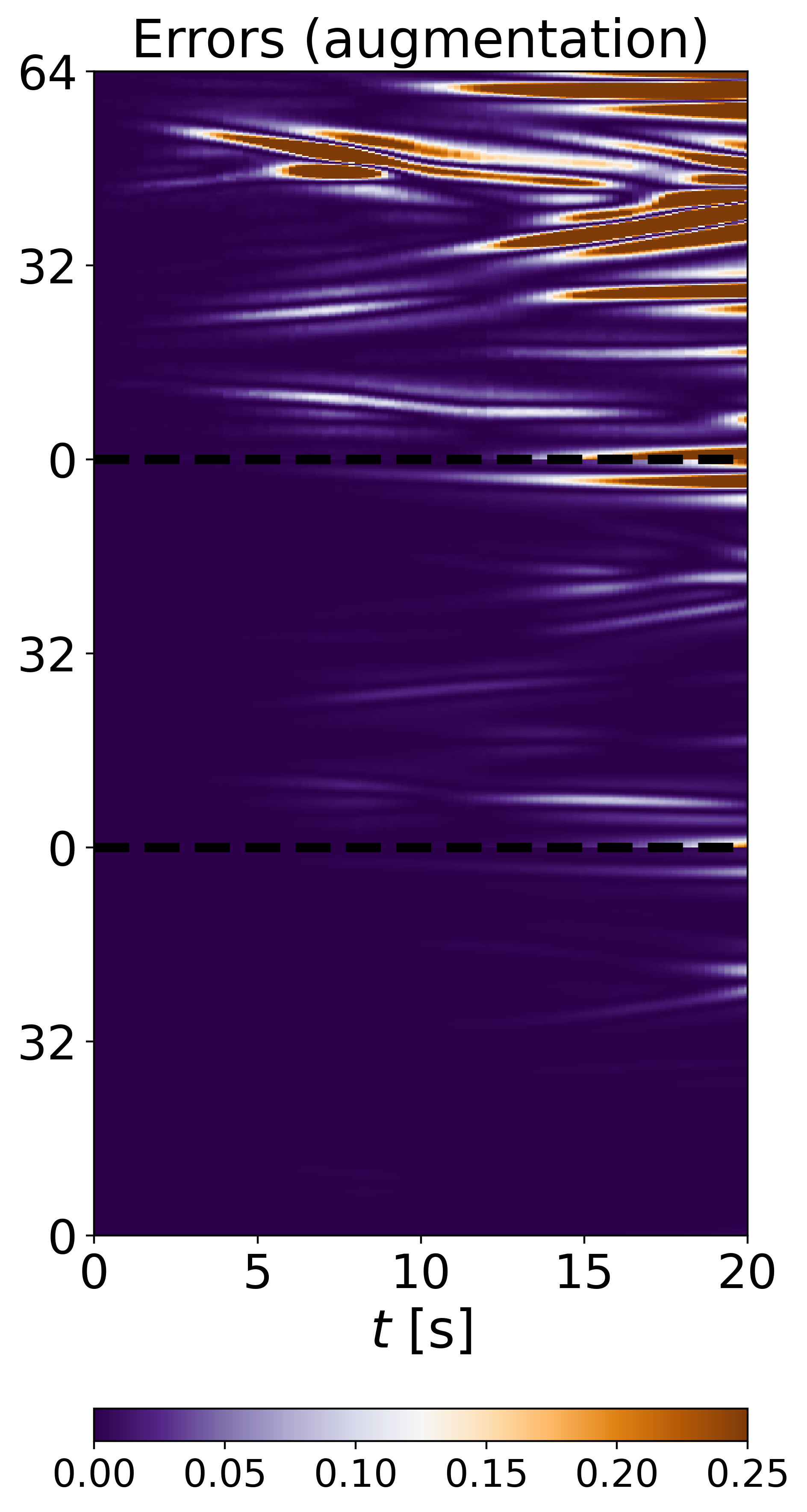}
    \includegraphics[width=0.325\linewidth]{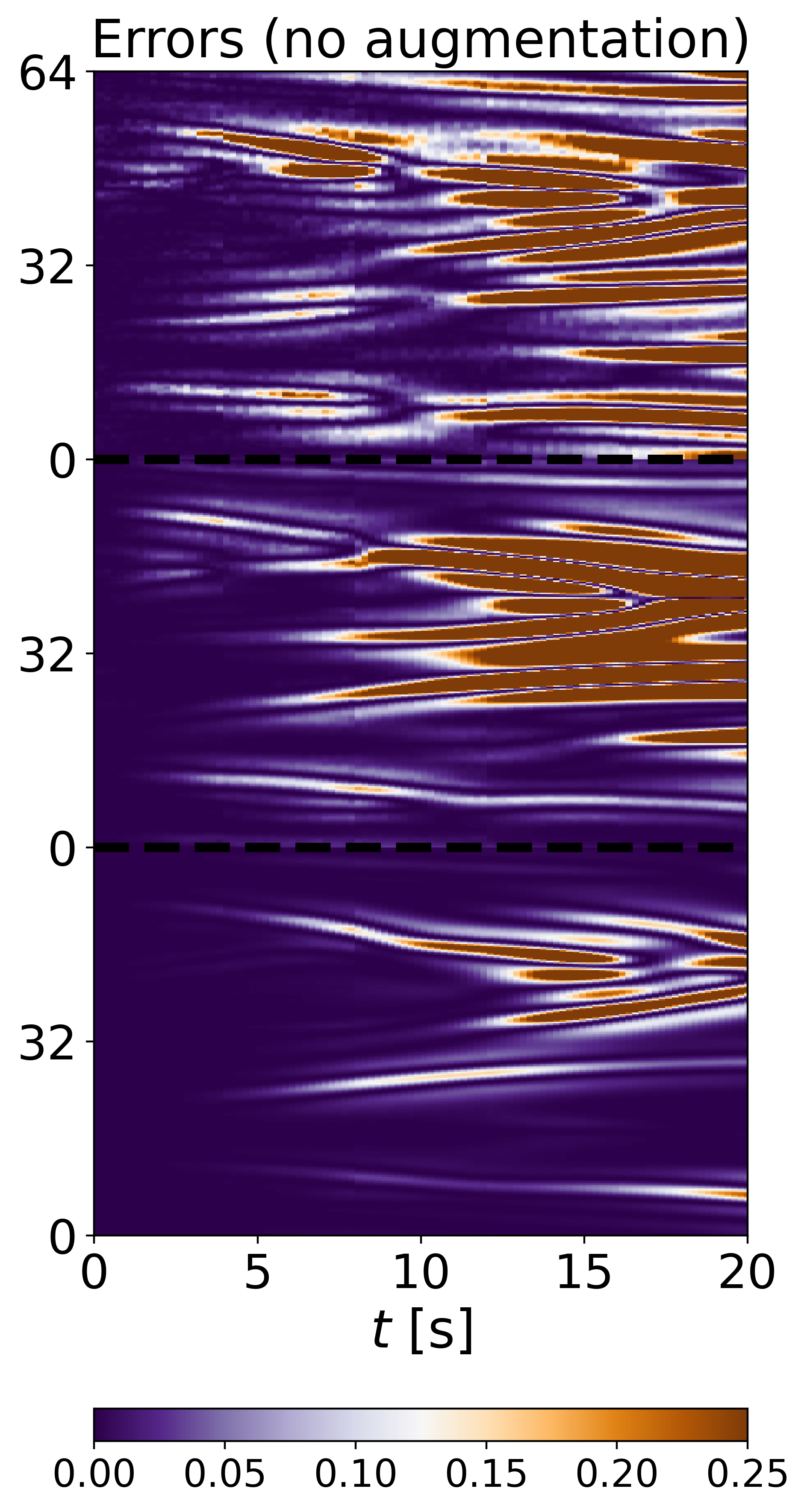}
    \caption{Visualisation of sample rollouts of the KS equation along with the corresponding per-cell normalized squared error.}
    \label{fig:KS_errors}
\end{figure}

\subsection{Different PDEs} \label{sec:pdes}
We tested LPSDA on the KdV, KS, and Burgers' equations. Each equation has symmetries listed in Table \ref{tab:symmetries}. Figure~\ref{fig:kdv-ks} shows sample complexity improvements on FNO, trained as a neural operator to predict $T \text{ s}$ into the future, as per Table~\ref{tab:settings}. This shows LPSDA improves sample complexity by $\times 4$ to $\times 16$. LPSDA works most strikingly on the Burgers' equation, where it appears the infinite-dimensional subalgebra $g_\alpha$ is doing most of the work. This symmetry takes two old solutions and combines them into a new one, in effect squaring the size of the dataset probably accounting for the near doubling in slope we observe (\textsf{Burgers' + G} line).

We plot on log-log axes and fit straight lines to the data, based on the documented phenomenon that neural network test performance follows a power law \citep{kaplan2020}. This experiment gives a strong indication that LPSDA could be effective for a great many other PDEs. Typically rollouts from the KS equation along with per-cell normalized squared errors is shown in Figure~\ref{fig:KS_errors}. Errors increase with time. LPSDA appears to delay the rate at which errors grow.

\subsection{Different models}  \label{sec:models}
Figure \ref{fig:kdv-model} shows the effect of LPSDA for the KdV equation. We compare the FNO and ResNet trained as neural operator methods. In development the ResNet struggled to model the solution manifold, so for a fairer comparison we made the problem easier by predicting $T=20 \text{ s}$ into the future instead of $T=40 \text{ s}$. Bold lines show test NMSE across training set size without LPDSA $(\emptyset)$ and dashed lines show it with LPSDA $(g_1g_2g_3g_4)$. We see that LPSDA is equivalent to increasing the training set size by a factor between $\times 4$ and $\times 8$. Again, we fitted straight lines because we are in the neural operator model setting. It is interesting that LPSDA improves each model by a differing amount, especially FNO, where the sample complexity improvement appears to grow with increasing dataset size.

\subsection{Different Training Methods}  \label{sec:setups}
Figure~\ref{fig:kdv-method} shows the LPSDA ablation for the FNO model trained either as a neural operator (blue), predicting $N_{t_\text{out}}=100$ timesteps in to the future in one shot, or as an autoregressive method (orange), predicting $N_{t_\text{AR}}=20$ timesteps at a time. Again LPSDA improves both methods. This experiment is important because rollout stability and accuracy of autoregressive methods---a still poorly understood phenomenon---appears to be intimately connected with generalization capacity \citep{um2021solverintheloop, BrandstetterEtAl2022}. LPSDA, in improving generalization performance, naturally improves AR methods as well as NO methods. It is interesting to note that the generalization performance of the AR method does not follow a power law.

\subsection{Incremental effect of symmetries} \label{sec:symmetry-ablation}
The effect of incrementally adding symmetries on the KdV-FNO neural operator model is shown in Figure~\ref{fig:kdv-multiple}. In Appendix \ref{sec:app:results} we show that this effect is repeated in other scenarios. We see that with this setup the time translation $(g_1)$ and Galilean boost $(g_3)$ improve sample complexity. Interestingly, space translation $(g_2)$ neither improves nor degrades performance. This is because the models we use are already translation equivariant. Furthermore, it appears scaling $(g_4)$ does not help either. Our best explanation is that with our parameterization scale is `too simple' a data transformation, simply multiplying solution values by a random constant. It is interesting to observe that not all symmetries improve performance equally; this corroborates what is commonly observed with classical data augmentation.
\begin{figure}[!b]
    \centering
    \includegraphics[width=0.495\linewidth]{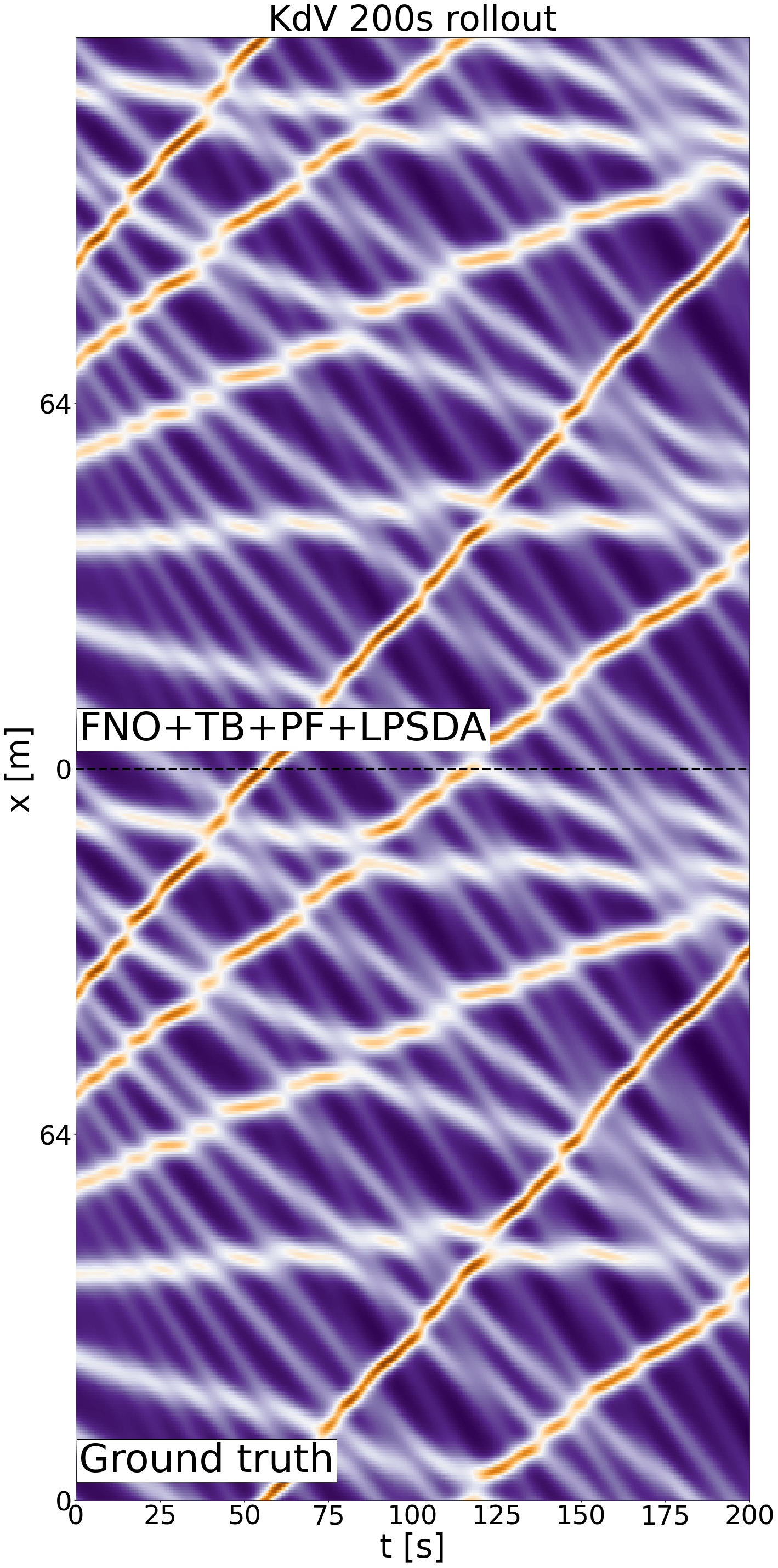}
    \includegraphics[width=0.495\linewidth]{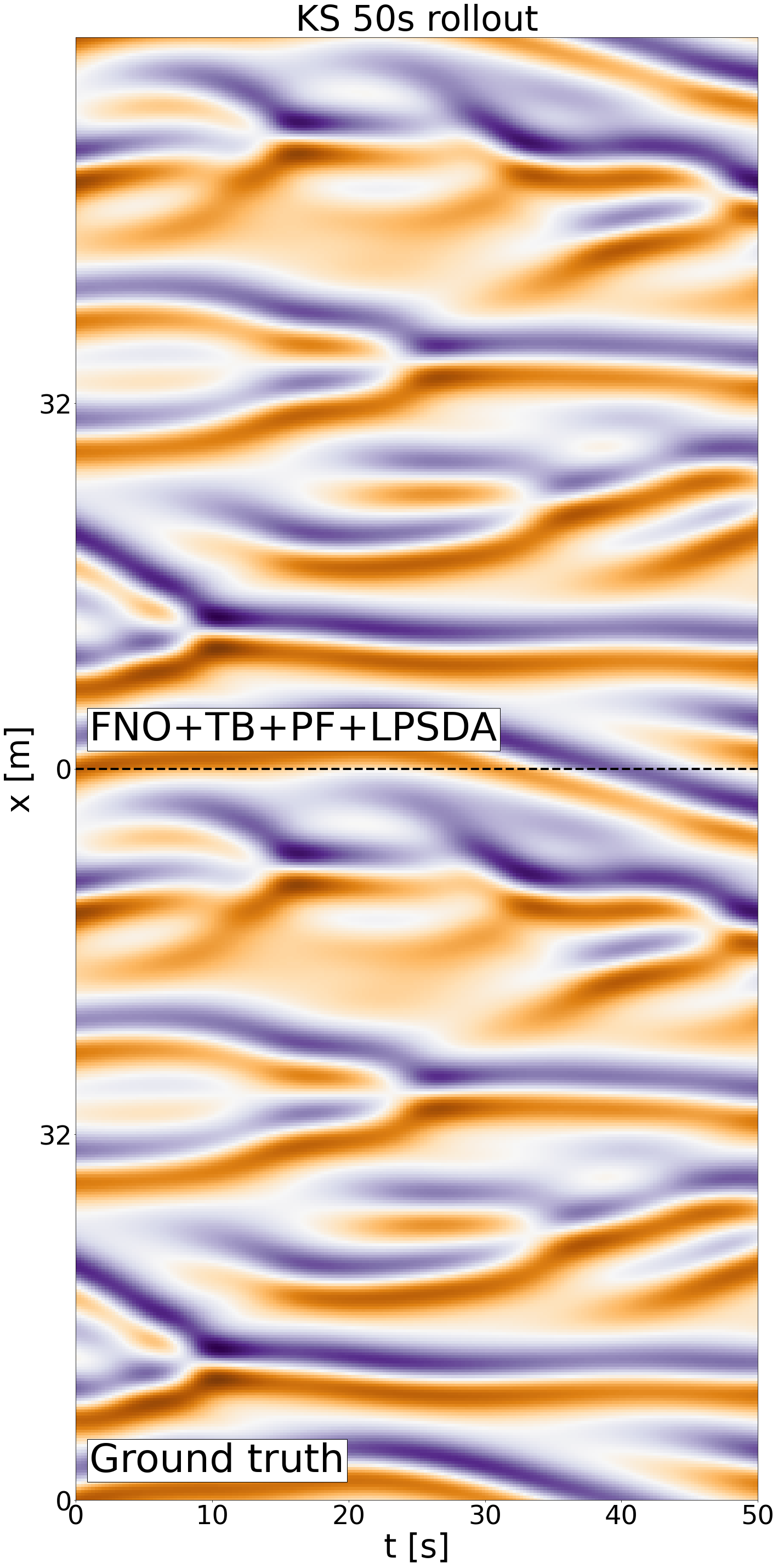} 
    \caption{Examples of ``long rollouts'' of $N_{t_\text{out}}=400$ timesteps on the KdV and KS equations. Visually it is hard to spot errors.}
    \label{fig:long_rollouts}
\end{figure}
\subsection{Very long rollouts} \label{sec:very-long}
\begin{figure*}[htb!]
    \centering
    \begin{subfigure}[b]{0.40\textwidth}
        \centering
        \includegraphics[width=\linewidth]{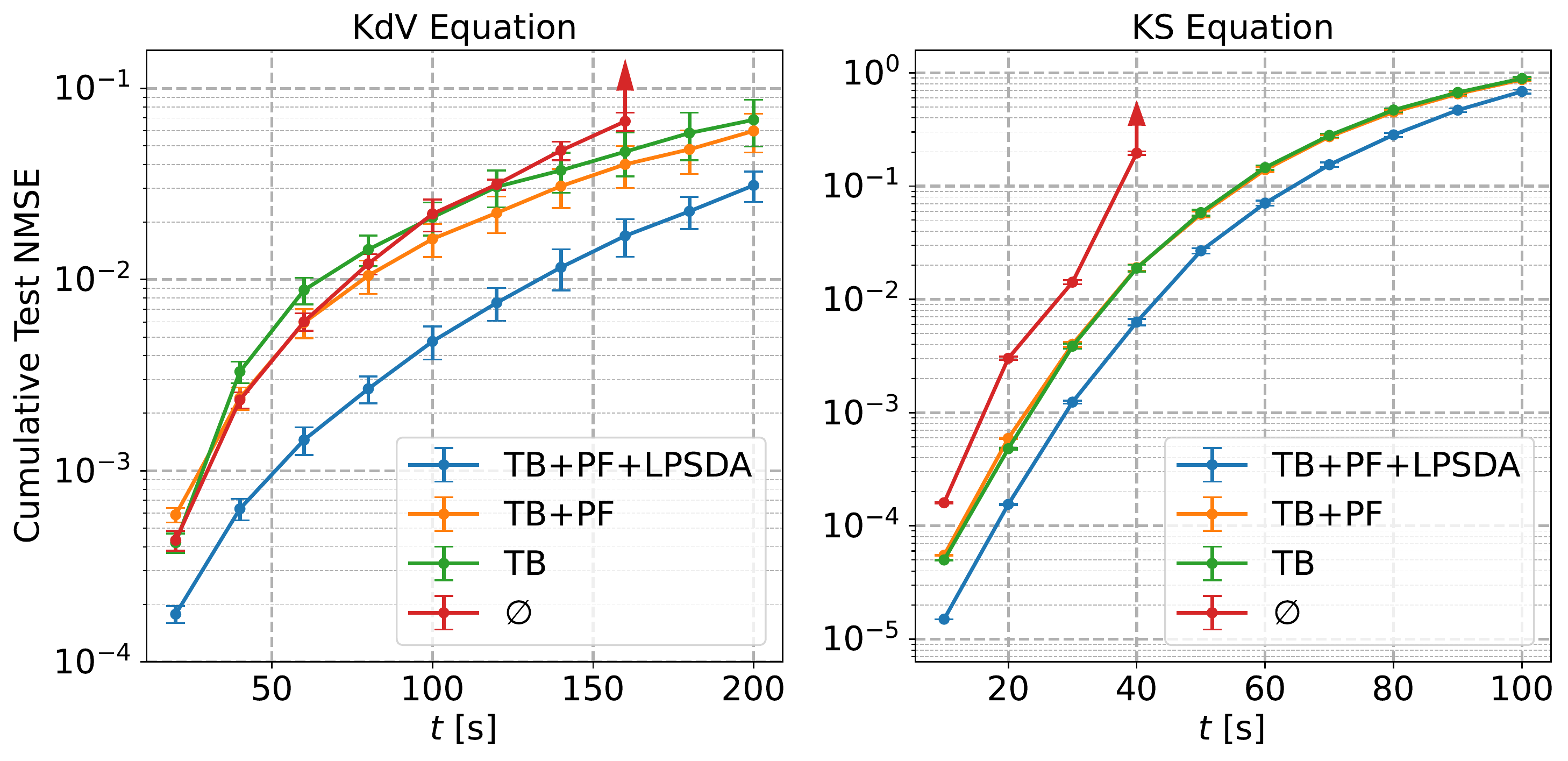}
        \caption{Long rollout errors}
        \label{fig:kdv-long}
    \end{subfigure}
    \begin{subfigure}[b]{0.29\textwidth}
        \centering
        \includegraphics[width=\linewidth]{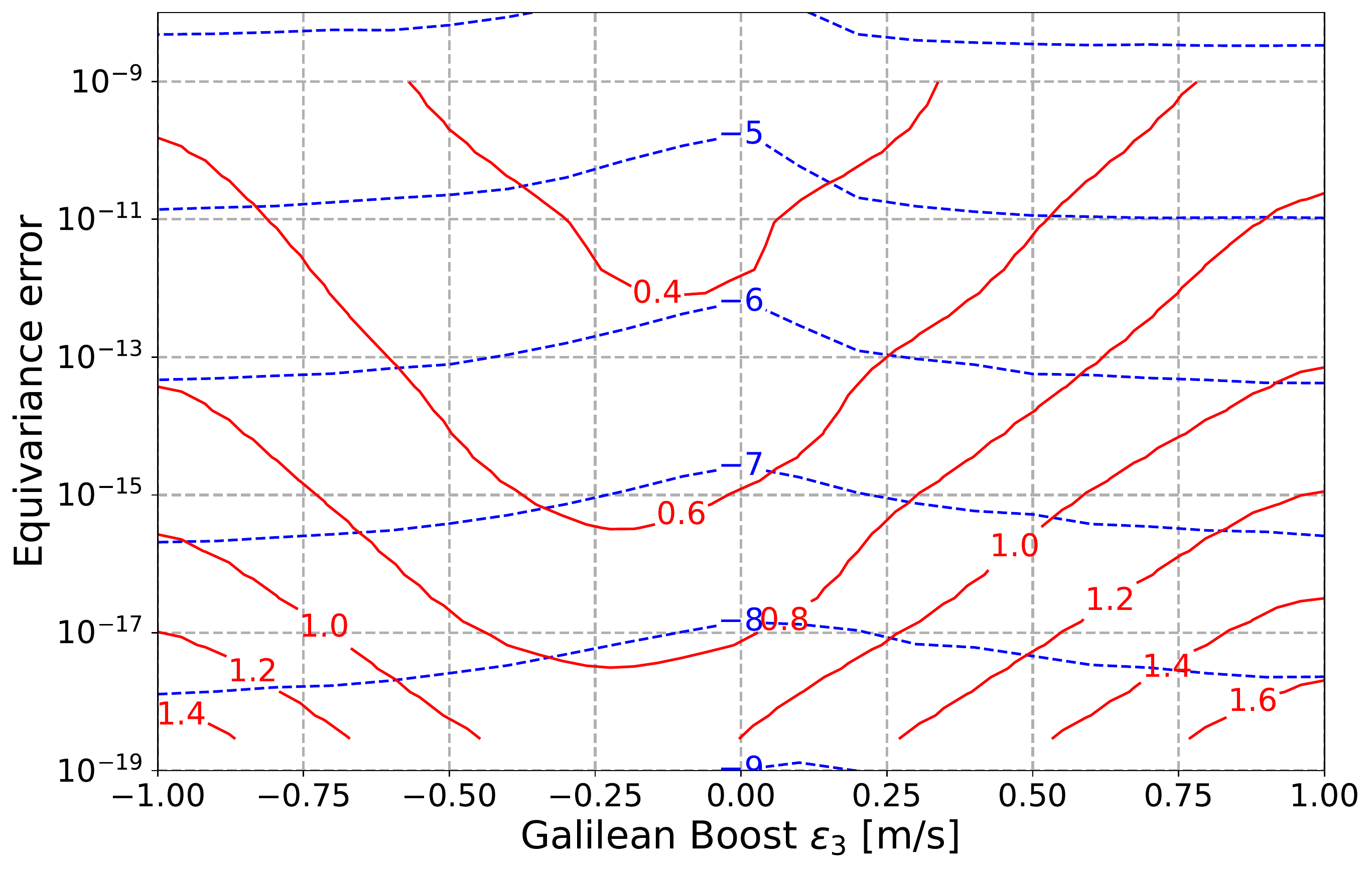}
        \caption{Classical equivariance analysis}
        \label{fig:CFL}
    \end{subfigure}
    \begin{subfigure}[b]{0.26\textwidth}
        \centering
        \includegraphics[width=\linewidth]{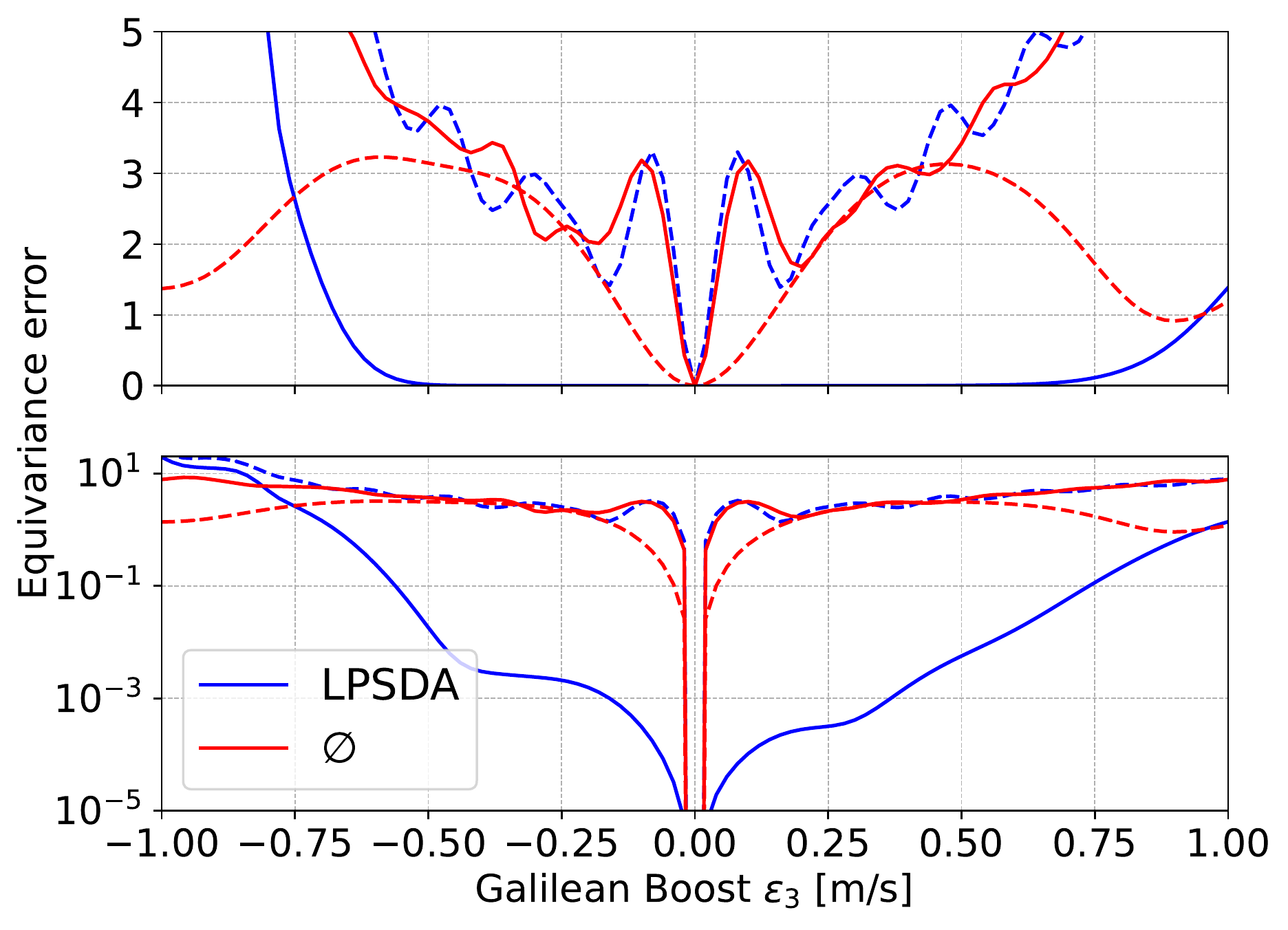}
        \caption{Neural equivariance analysis}
        \label{fig:equi-error}
    \end{subfigure}
    \caption{(a) Cumulative error growth over a long autoregressive rollout ($N_{t_\text{out}}=400$ timesteps) on the KdV and KS equations. Uparrows indicate error explosion upon rollout instability. The four incrementally combined settings are: $\emptyset$, one timestep prediction at a time; \textsf{TB}, temporal bundling \citep{BrandstetterEtAl2022}; \textsf{PF}, the pushforward trick; \textsf{LPSDA}, our method. (b) Equivariance error of a single KdV rollout after $T=100 \text{s}$ against Galilean boost parameter $\eps_3$. We use $\texttt{tol} = \texttt{1e-12}$ as groundtruth. {\color{blue}\textsc{Blue dashed}}: Isotherms of log solver tolerance (base 10). Equivariance error decreases with log tolerance, relatively unaffected by transformation magnitude. {\color{red}\textsc{Red solid}}: Isotherms of solution time. Solution time is a function of tolerance and $\eps_3$. Neural solvers instead have constant time overhead. (c) Equivariance error of neural models trained with (\textsf{LPSDA}) and without LPSDA $(\emptyset)$. Bold lines: Equivariance error $\|f(g\rvu_0) - gf(\rvu_0)\|_2^2 / \|f(g\rvu_0)\|_2^2$ (where $f$ is model output). \textsf{LPSDA} is clearly equivariant around the training range $\eps_3\in[-0.4, 0.4]$ and $\emptyset$ is not. Dashed lines: Autocorrelative error $\|f(g\rvu_0) - f(\rvu_0)\|_2^2 / \|f(\rvu_0)\|_2^2$. $\emptyset$ maps $f(g\rvu_0)$ close to $f(\rvu_0)$ for small $\eps_3$ but \textsf{LPSDA} does not. We suspect $\emptyset$ tries to map unseen $\eps_3$ back into a region ($\eps_3=0$) often encountered in the training set.}
\end{figure*}

We ran long rollouts on the KdV (KS) equation of $T=200\text{ s}$ ($T=50\text{ s}$) with $N_{t_\text{out}}=400$ timesteps to check how LPSDA can improve performance in a practical use case. We used an AR-trained FNO model. We compare a na\"ive model predicting $N_{t_\text{AR}}=1$ timestep at a time $(\emptyset)$, a temporally bundled model predicting $N_{t_\text{AR}}=120$ timesteps at a time (\textsf{TB}), the $N_{t_\text{AR}}=120$ model with the pushforward trick \citep{BrandstetterEtAl2022} (\textsf{TB+PF}) needed for long run stability, and (\textsf{TB+PF+LPDSA}) where we add LPSDA. Figure \ref{fig:long_rollouts} shows the resulting rollouts and Figure \ref{fig:kdv-long} shows the cumulative NMSE with respect to time. Visually it is difficult to spot errors. The error graphs, on log-lin plots, reveal interesting subexponential cumulative error growth with time. The uparrows for the $\emptyset$ rollouts indicate error explosion. Here, again, we see that LPSDA improves rollout performance. A summary of all results is in Appendix \ref{sec:app:results}.  

\subsection{How equivariant are classical/neural solvers} \label{sec:equivariance-error}
Figure~\ref{fig:CFL} shows the equivariance properties of classical solvers for KdV Galilean boost parameter $\eps_3$ after $T=100 \text{ s}$. For model output $\rvu$ at time $T$, this is $\|\rvu(g\rvu_0) - g\rvu(\rvu_0)\|_2^2 / \|\rvu(g\rvu_0)\|_2^2$. Ground truth was solved at tolerance $\texttt{1e-12}$. Blue dashed lines are isotherms of constant log solver tolerance (base 10). This error is mainly a function of solver tolerance. The red solid lines are isotherms of constant solution time. Thus more time is needed to maintain a constant error for larger $|\eps_3|$. Neural solvers, on the other hand, have constant time overhead (our AR-FNO model measured $38$ ms), but without error control. Figure~\ref{fig:equi-error} shows equivariance error of neural models trained with and without LPSDA. LPSDA clearly encourages equivariance around the training range $\eps_3\in[-0.4, 0.4]$. Classical solvers have superior equivariance error are two orders slower. We also considered autocorrelative error defined as $\|\rvu(g\rvu_0) - \rvu(\rvu_0)\|_2^2 / \|\rvu(\rvu_0)\|_2^2$. Models trained without LPSDA map $\rvu(g\rvu_0)$ close to $\rvu(\rvu_0)$ for small $\eps_3$. We suspect the $\emptyset$ model outputs solutions close to regions of the solution manifold it has observed, where $\eps_3 \simeq 0$. This shows how data augmentation is important for generalization.

\paragraph{Speed and Difficulty.}
We highlight the $\eps_3$ dependence of the time isotherms in Figure~\ref{fig:CFL}. Adaptive timestep solvers limit the error incurred across a given unit of time by expending more operations on more `difficult' trajectories. Consider a wave moving across a discrete spatial grid. To compute its position at a future timestep, we are limited by the maximum speed information can propagate across the grid. This must exceed the speed of the wavefront, otherwise we have to reduce the time increment. Faster waves require finer timesteps. For the KdV equation, we can control wavespeed by adding $\eps_3$ to the initial conditions. Larger $|\eps_3|$ leads to faster waves, and thus longer solver times. By contrast, data augmentation is several orders of magnitude faster (our implementation: $0.76$ ms). This indicates training data generation could be sped up by solving `simple' problems on a classical solver, and then transforming solutions into challenging regions of the training data space.

\section{Related Work} \label{sec:related-work}
The closest work to ours is \citet{WangWY21} who design neural PDE solvers equivariant to Lie point symmetries of the Navier-Stokes equation and Heat equation. While this method is sensible and effective, they are only able to incorporate a single symmetry (other than translation and time) at a time. With our method, we can learn equivariance to multiple symmetries at once. It is also simpler to implement a data augmentation pipeline than a equivariant network. Furthermore, the computational budget for equivariant networks is exponential in the number of symmetries, a limitation if you have 6 or 7 symmetries, a scenario unseen in the equivariance literature \citep{bronstein2021}.

In the context of classical PDE solvers \citet{Hoarau2007a, Hoarau2007b} analyzed the stability properties of various finite difference schemes under the action of the individual symmetries of the Burgers' equation. Studying equivariance properties of classical solvers dates back to 1994 with \citet{Dorodnitsyn1994FiniteDM}. \citet{Valiquette_2005} discretize scalar-valued PDEs with two independent variables such that their Lie point symmetries are preserved.

Symplectic integrators \citep[\S~II.16]{HairerNH93}, such as leapfroging \citep[p.~72]{Neal93probabilisticinference}, have a long history in machine learning for their utility in Hamiltonian Monte Carlo \citep{duane1987} and \citep[\S~5]{Neal93probabilisticinference}. These are limited to integrating symplectic systems, such as Hamiltonians, which abound in physical systems. More recent neural models learn Hamiltonians directly \citep{GreydanusEtAl2019} or indirectly \citep{Sanchez2019, Cranmer2020, FinziWW20}. \citet{Bar_Sinai_2019, KochkovEtAl2021} explicitly conserve volume (another integral of motion) for conservation form evolution equations via finite volume projection on their solver outputs. 

In the field of neural PDE solvers, there are neural operator methods of finite dimensionality \citep{Raissi18, SirignanoS18, Bhatnagar2019, Guo2016, Zhu2018, Khoo2020} and infinite dimensionality \citep{li2020fourier, bhattacharya2020model, Patel2021}. Leading autoregressive methods are \citet{Bar_Sinai_2019, um2021solverintheloop, SanchezEtAl2020, BrandstetterEtAl2022}. Then in so-called \emph{neural augmentation} a neural component is added to finite elements \citep{HsiehZEME19}, multigrid solvers \citep{GreenfeldGBYK19}, and eikonal solvers \citep{LichtensteinPK19}.

\section{Limitations and Restrictions} \label{sec:limitations}
The space of PDEs and their symmetries is vast. We made some experimental design choices to limit the scope of our investigations. Namely, we investigated Eulerian solvers (instead of Lagrangian) on 1D evolution equations on periodic domains. These choices are a restriction of our experimental setup and not of the core idea. From a theory standpoint we expect LSPDA to generalize easily to other settings.

Methodologically, we could only implement symmetries that act globally. Purely local symmetries were not tackled, which would require data augmentation on local patches, something we are yet to develop. Furthermore, while theory provides an exhaustive list of Lie point symmetries for a given PDE, it gives no indication as to the efficacy of each symmetry. We could reason qualitatively about efficacy for, say, space translation or the infinite-dimensional subalgebra $g_\alpha$, but we could not provide quantitative predictions. Lastly, given a PDE, user input is required to implement the data augmentation functions from a mathematical description of the symmetries $\{g_1, ..., g_d\}$. Mechanizing this process would make LPSDA far more attractive. This said, we feel LPSDA is a significant step forward in terms of expanding the range of techniques available to train neural PDE solvers.

\section{Conclusion and Future work} \label{sec:conclusion}
We showed it is possible to apply the commonplace practice of data augmentation to the training of neural PDE solvers. Here, the space of possible transformations, restricted to Lie point symmetries, can be exhaustively derived from the PDE definition. This places a typically intuition-based aspect of the deep learning user experience on firm mathematical footing. We demonstrated our claims on three common fluid dynamical PDEs: the Korteweg-de Vries, Kuramoto-Shivashinsky, and Burgers' equations. Here we showed clear decrease in sample complexity, measured by generalization error across PDEs, models, and training regime. We investigated the effect of individual symmetries and the equivariance properties of solvers, classical and learned.

There are many potential directions from here; symmetries beyond the Lie point symmetries are vast. Examples include \emph{generalized symmetries}, such as \emph{contact symmetries} and \emph{Lie-B\"acklund symmetries} (c.f.,\ Section~\ref{sec:background}). These form a broader class of transformations, depending on solution derivatives as well as value. For instance, the KdV equation has an infinite number of these symmetries. Then there are discrete symmetries, such as reflections, which can easily be integrated into our framework. If a problem can be cast into variational form, then Noether's theorem states that every symmetry of the variational form translates into a conserved quantity of the system. This has been exploited by the ML4Physics community with Lagrangian Neural Networks~\citep{GreydanusEtAl2019}, and with classical solvers this is the basis of the finite volume method and symplectic integrators.

Orthogonally, we could follow the work pioneered by \citet{WangWY21} and bake equivariance into the solver architecture itself. Beyond ``simple'' symmetries, such as translation or scale (which is actually very difficult), there is little work in this area, apart from \citet{finzi2020generalizing}. In Section~\ref{sec:equivariance-error}, we saw that classical solvers can exhibit controllable equivariance error. We speculate that merging neural PDE solvers with variable tolerance solvers may be a fruitful area for building equivariance to complicated groups.

\section{Acknowledgements}
The authors thank Markus Holzleitner for discussions on prolongation theory, and especially for pointing us towards the Cauchy-Kovalevskaya theorem.

\bibliography{main}
\bibliographystyle{icml2022}

\newpage

\appendix

\section{Training data} \label{sec:data}

\paragraph{Korteweg-de Vries and Kuramoto-Shivashinsky equation.}
Numerical groundtruth data for the Korteweg-de Vries (KdV \citep[p.~360]{Boussinesq}\citep{KdV} and the Kuramoto-Shivashinsky (KS) \citep{Kuramoto1977, Shivashinsky1977} equation is obtained on a periodic domain using the \emph{method of lines}, with the spatial derivatives computed using the pseudo-spectral method. 
The \emph{method of lines} discretizes the domain $\gX$ on a (regular) mesh at a finite number of points $X = \{x_i \in \gX\}$, where $|X| < \infty$. We can then represent the signal $\{u(t, x) | x \in X, t \in [0,T] \}$ in terms of a vector $\rvu(t)$, where the $i$\textsuperscript{th} element of the vector is the $i$\textsuperscript{th} element in $X$ (assuming we have imposed some sensible ordering on $X$). We then solve
\begin{align}
    \partial_t \rvu = f(t, \rvu) \qquad t \in [0,T] \ , \label{eq:coupled-odes}
\end{align}
where $f$ is a numerical implementation of all the spatial derivatives. This is just a set of coupled ODEs. Typically we also discretize the time domain $[0,T]$ as well to yield a forward time-stepping formulation.
In the pseudospectral method, the derivatives are computed in the frequency domain by first applying a fast fourier transform (FFT) to the data, then multiplying by the appropriate values and converting back to the spatial domain with the inverse FFT. This method of differentiation is implemented by the \texttt{diff} function in the module \texttt{scipy.fftpack}.
For integration in time we use an implicit Runge-Kutta method of Radau IIA family, order 5~\citep{HairerNH93}.

As a crosscheck, we use a Finite Volume method (FVM),
which is a valid approach since the equations at hand are in conservation form, i.e.
it can be shown via the divergence theorem that the integral of $\rvu$ over cell $i$ increases only by the net flux into the cell. We therefore estimate the flux at the left and right cell boundary at time $t_k$, which we again do via the pseudospectral method of the module \texttt{scipy.fftpack}.

\begin{figure*}[htb!]
    \centering
    \begin{subfigure}[b]{0.23\textwidth}
        \centering
        \includegraphics[width=\textwidth]{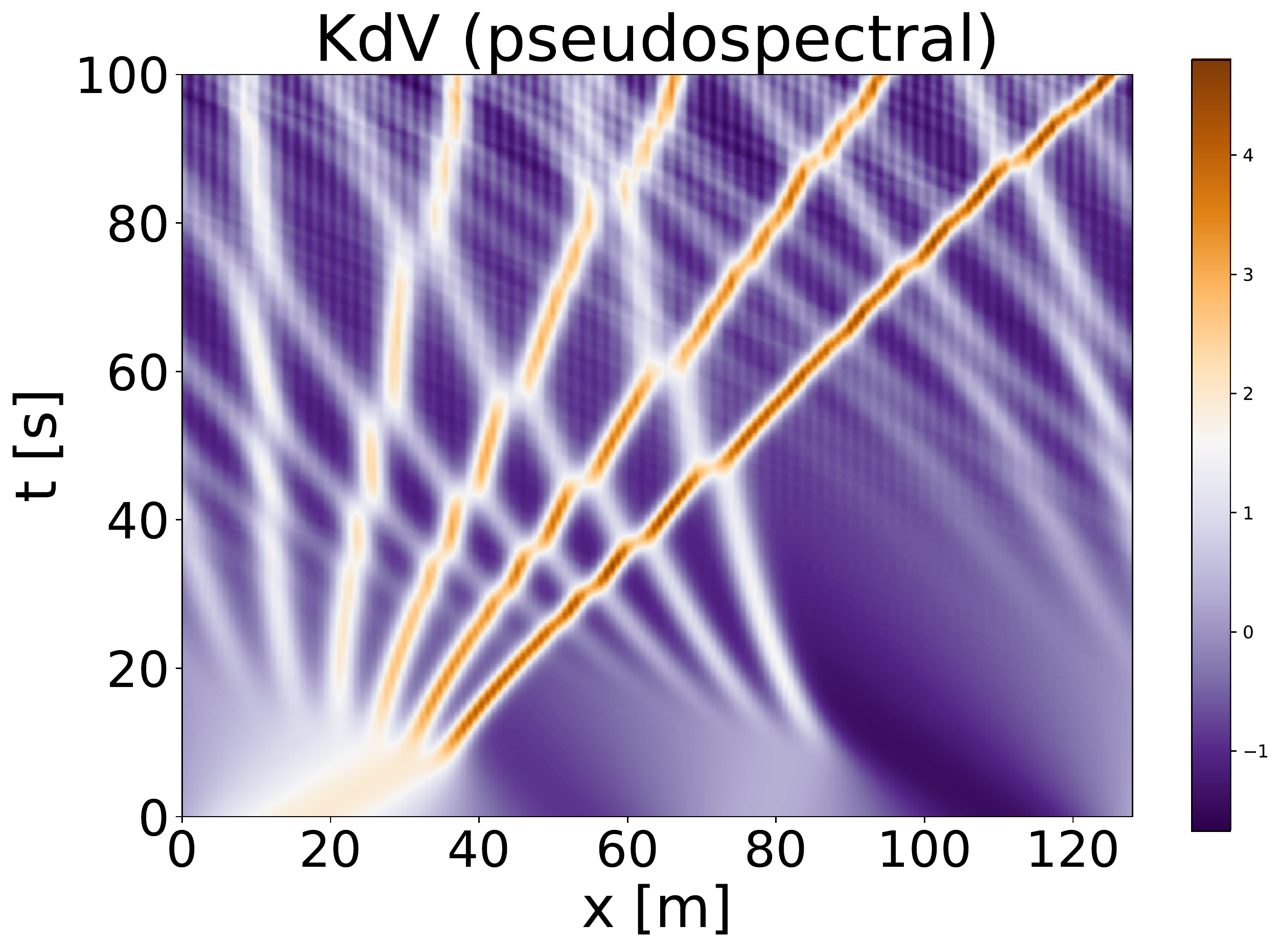}
        \caption{}
    \end{subfigure}
    ~
    \begin{subfigure}[b]{0.23\textwidth}
        \centering
        \includegraphics[width=\textwidth]{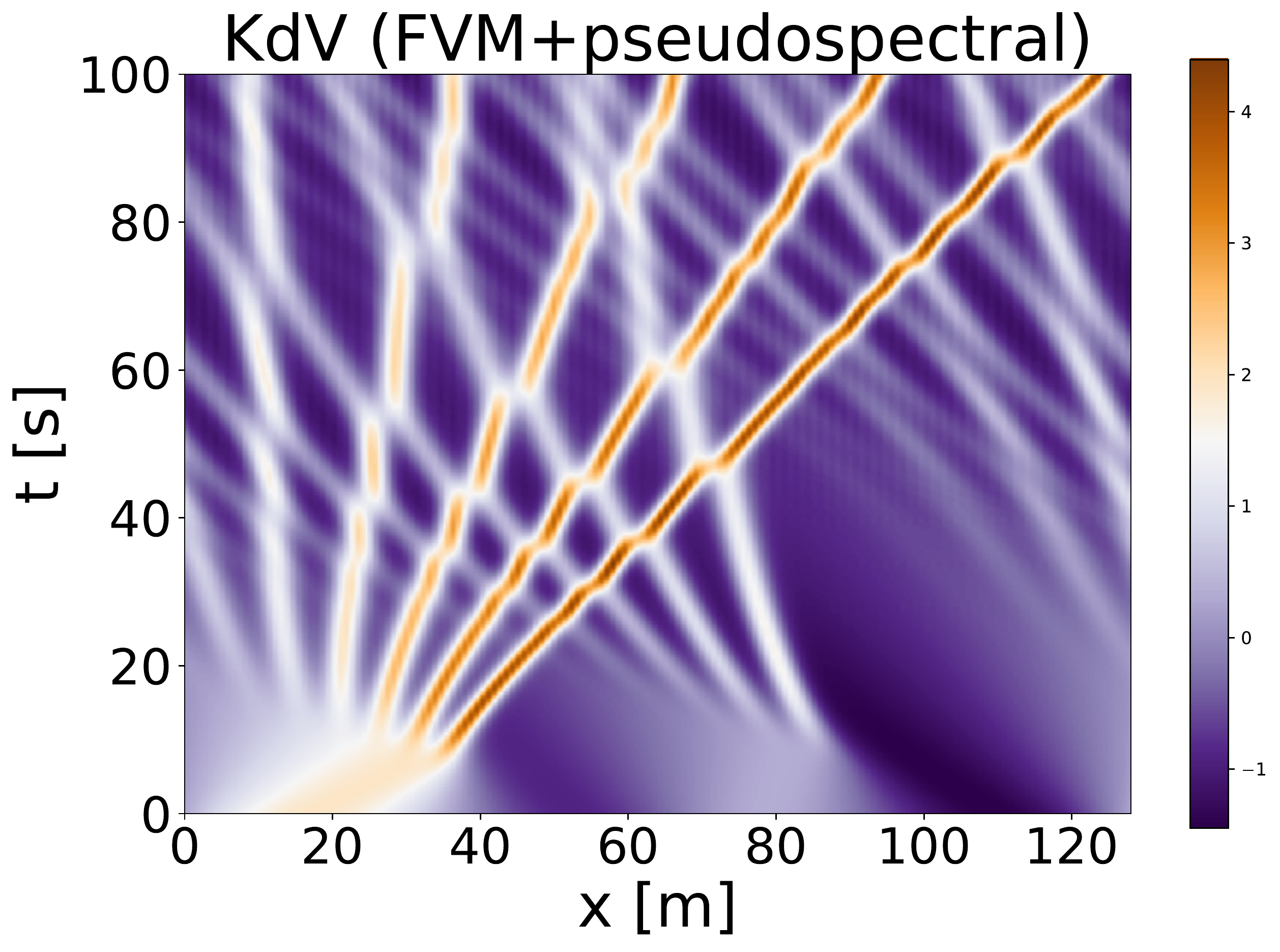}
        \caption{}
    \end{subfigure}
    ~
    \begin{subfigure}[b]{0.23\textwidth}
        \centering
        \includegraphics[width=\textwidth]{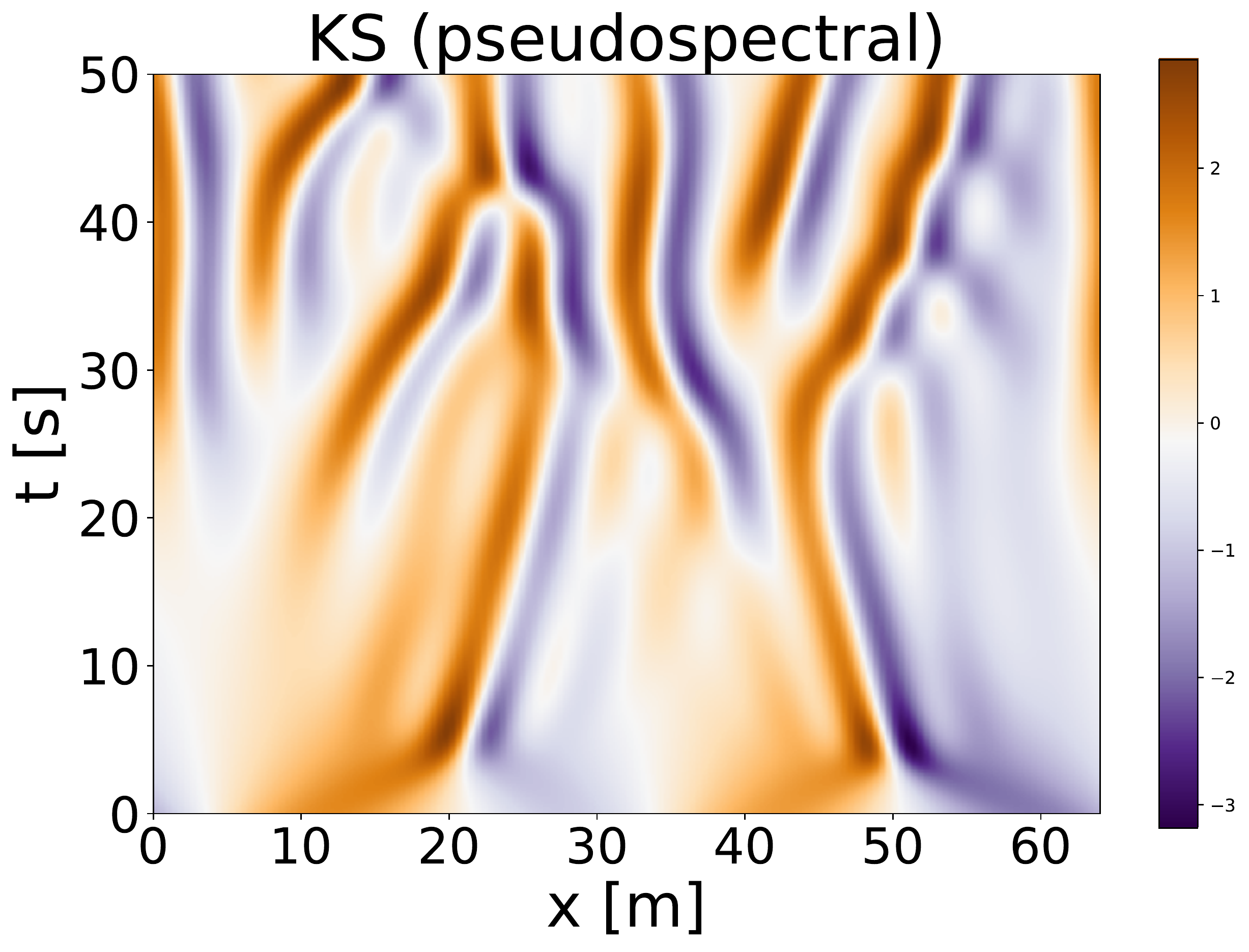}
        \caption{}
    \end{subfigure}
    ~
    \begin{subfigure}[b]{0.23\textwidth}
        \centering
        \includegraphics[width=\linewidth]{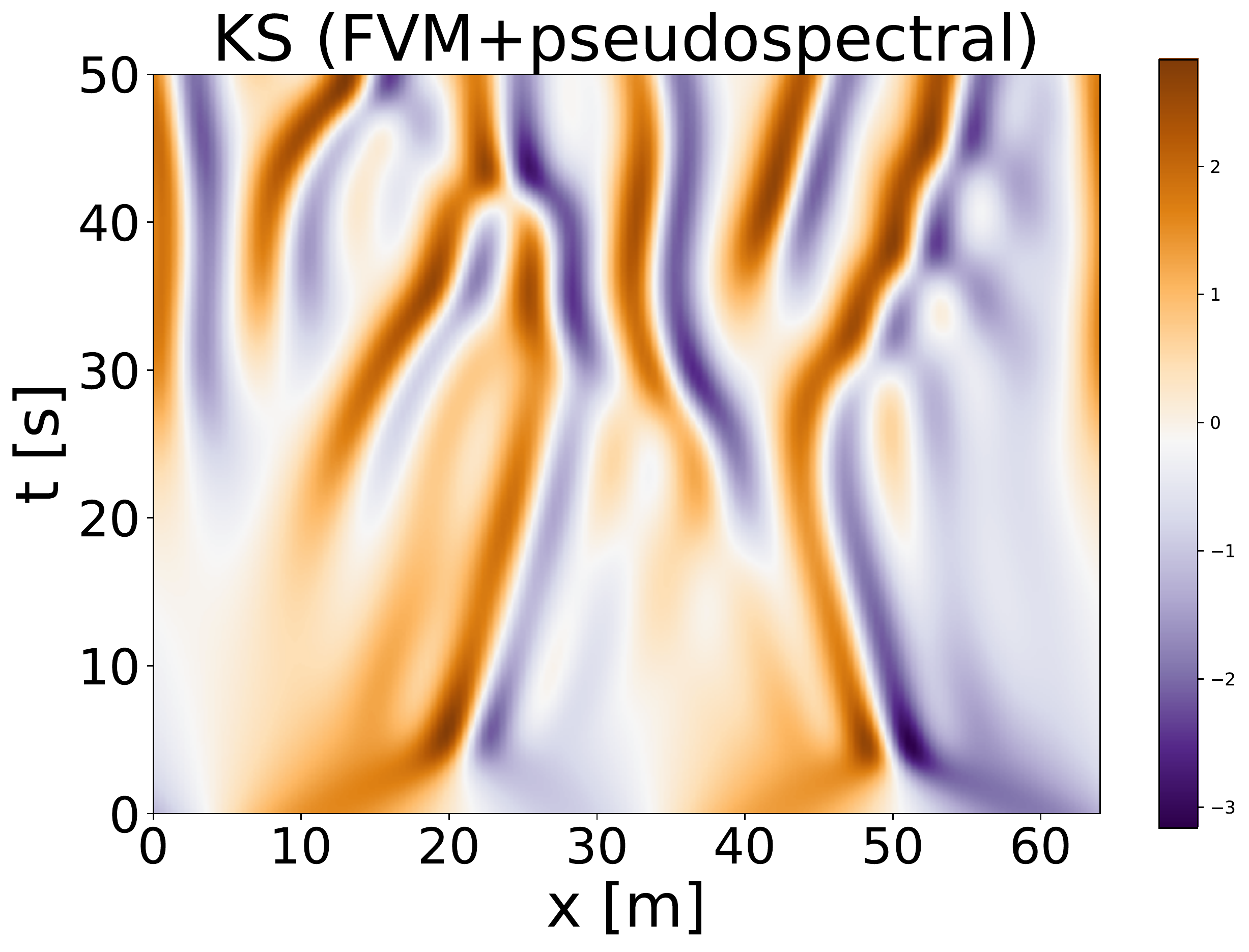}
        \caption{}
    \end{subfigure}
    \caption{Comparison of pseudospectral and Finite Volume Method (FVM) pseudospectral data generation for the Korteweg de Vries (a,b) and the Kuramoto-Shivashinsky (c,d) equation. Mean numerical mean-squared errors are in the order $10^{-5}$.}
    \label{fig:data_generation}
\end{figure*}

Figure \ref{fig:data_generation} shows visual comparisons of the described pseudospectral and FVM pseudospectral data generation. Mean numerical mean-squared errors are in the order of $10^{-5}$.

\begin{figure}[!htb]
    \centering
    \includegraphics[width=\linewidth]{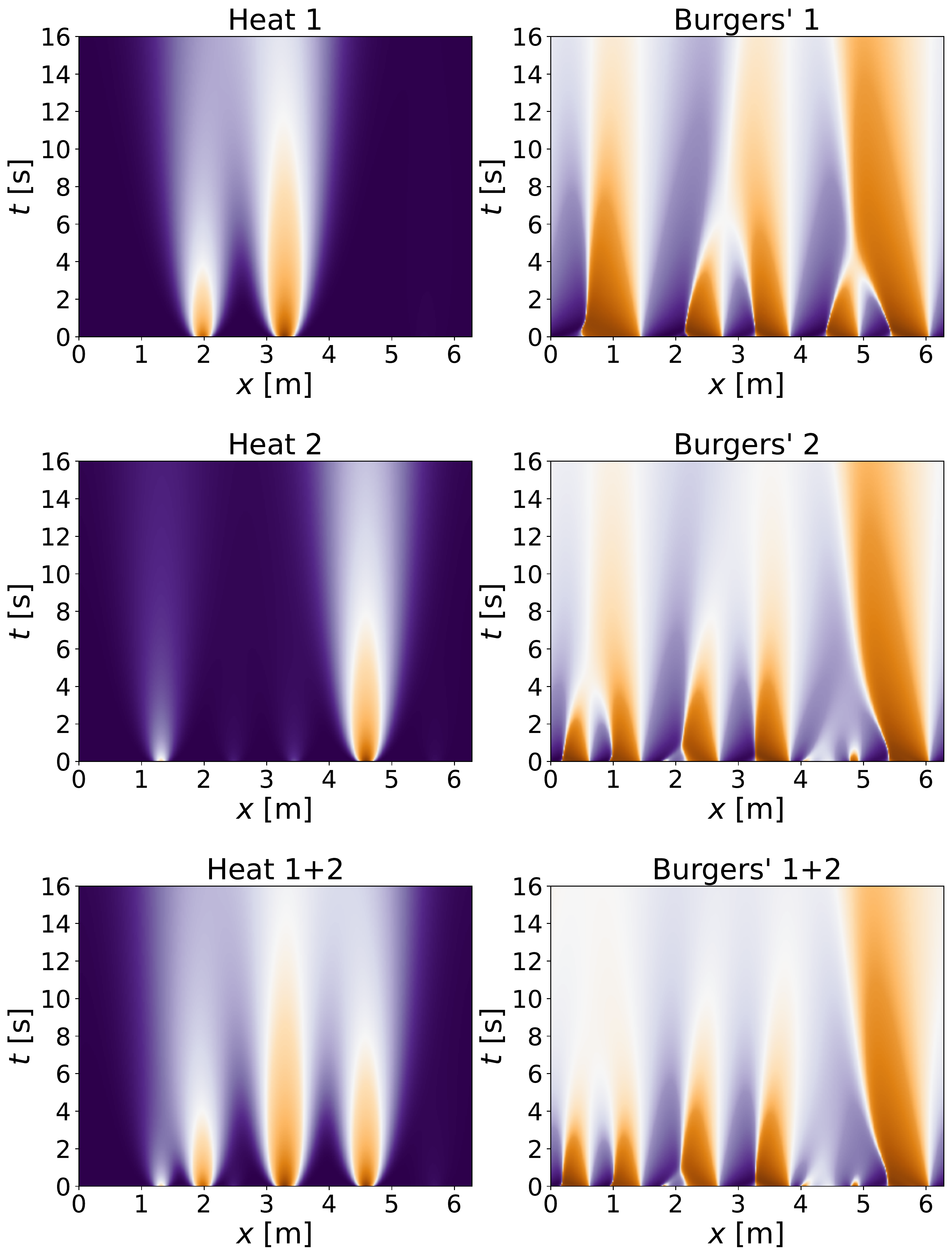}
    \caption{Comparison of exemplary trajectories of the Heat and the Burgers' equation. The Burgers' equation is obtained out of the Heat equation via the inverse Cole-Hopf transformation.}
    \label{fig:data_generation_Burgers}
\end{figure}

\paragraph{Burgers' equation.}
Numerical groundtruth data for the Burgers' equation \citep{Burgers1948}  
\begin{align}
    u_t + uu_x - \nu u_{xx} = 0 \ ,
\end{align}
is obtained by solving the Heat equation 
for viscocity $\nu = 0.01$.
This is possible due to the Cole-Hopf transformation~\citep{Cole1951, Hopf}
\begin{align}
    u = 2\nu \frac{\partial}{\partial x} \log(\phi) \ ,
\end{align}
which turns the Burgers' equation into the Heat equation:
\begin{align}
    \phi_t = \nu \phi_{xx} \ .
\end{align}
The Heat equation is solved in the same manner as the KdV and the KS equations, initial conditions for the Heat equation are obtained by taking initial conditions of the Burgers' equation and applying an inverse Cole-Hopf transformation.
Finally, applying the inverse Cole-Hopf transformation on the solved trajectory yields the Burgers' equation. Figure \ref{fig:data_generation_Burgers} shows exemplary Heat equation and respective Burgers' equation trajectories. The trajectories \textit{Burgers' 1+2} are of special interest since they are obtained via non-linear combination of the trajectories \textit{Burgers' 1} and \textit{Burgers' 2}, which is exactly what the generator $g_{\alpha}$ in Table \ref{tab:symmetries} stands for.

\section{Interpolation scheme} \label{sec:interpolation}
In our method we use trigonometric interpolation to shift/rotate solutions in the $x$-direction. This makes use of the Fourier shift theorem: translating a function $u(x) \mapsto u(x - \eps)$ on periodic domain of length $L$ linearly shifts its Fourier transform as $\hat{u}(k) \mapsto e^{-i 2 \pi k \eps / L } \hat{u}(k)$. We can thus translate a function by $\eps$ by first taking a Fourier transform $\gF$, shifting the phase and then taking the inverse Fourier transform $\gF$:
\begin{align}
    u(x) \overset{\gF}{\mapsto} \hat{u}(k) \overset{\text{shift}}{\mapsto} e^{-i 2 \pi k \eps / L }\hat{u}(k) \overset{\gF^{-1}}{\mapsto} u(x - \eps) \ ,
\end{align}
We originally chose this scheme based on standard signal processing considerations. Furthermore, it seemed a sensible choice, given that we use pseudospectral methods to compute the instantaneous spatial derivatives in the classical solver. 

\begin{figure}[!htb]
    \centering
    \includegraphics[width=\linewidth]{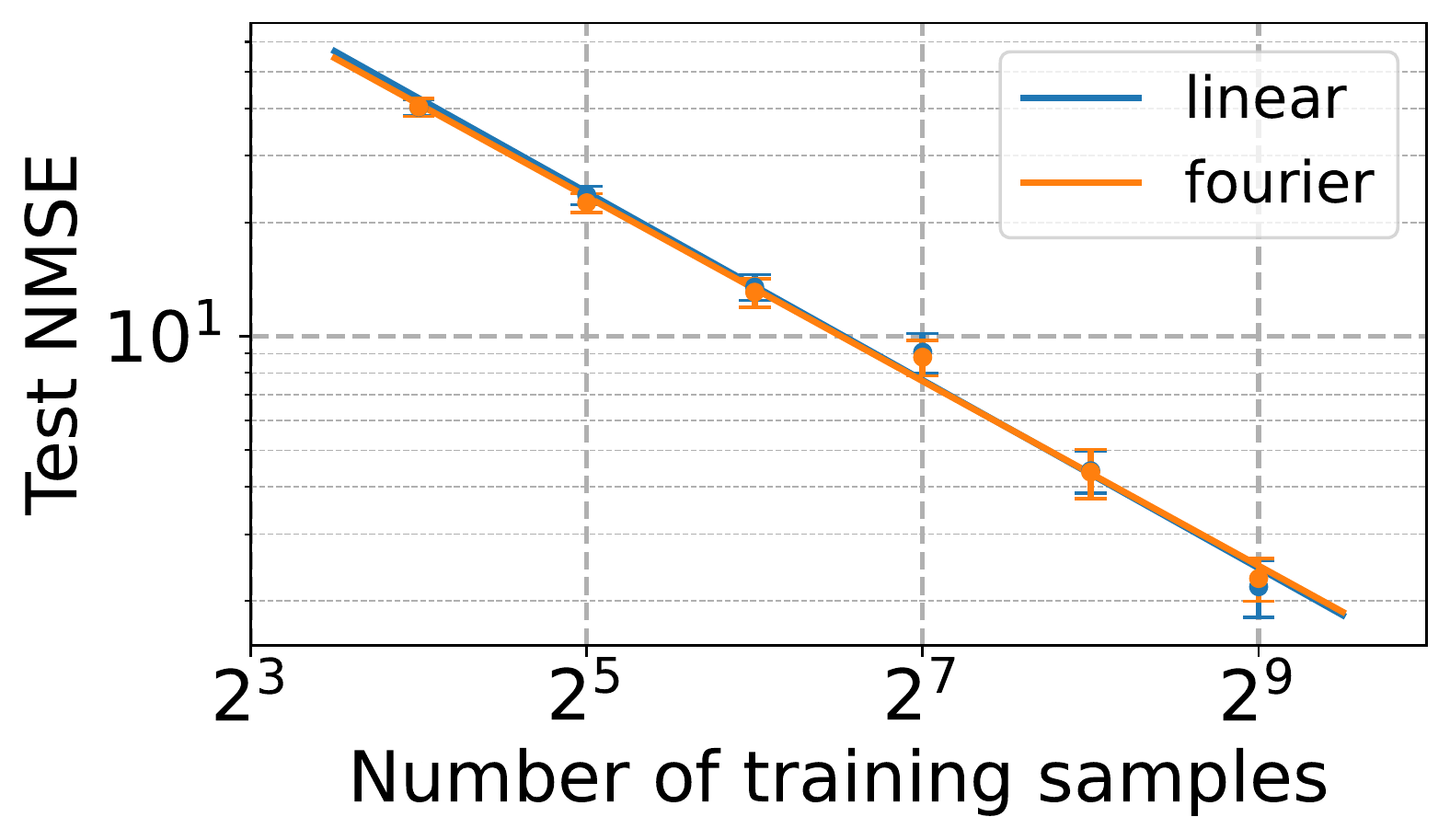}
    \caption{Comparison of the linear and Fourier interpolations schemes for different training set sizes, over the KdV equation, using the setup from Section~\ref{sec:pdes}. We see no discernable difference in performance.}
    \label{fig:linear-fourier}
\end{figure}

We benchmarked the performance of Fourier interpolation against linear interpolation after experiments as a crosscheck. The results are in Figure~\ref{fig:linear-fourier}. Surprisingly, the results show that whether we interpolate data using trigonometric or linear interpolation the effect on test NMSE is not statistically significant, contrary to what we primarily had suspected. This result does not invalidate our experiments. We suspect that the neural models are able to undo interpolation artifacts introduced by the linear interpolation scheme. This is potentially because the neural models take in not just an initial condition at a single time, but a trajectory of the previous $N_{\text{in}}$ steps, across which they can average out interpolation artifacts.

\section{Experimental Architectures} \label{sec:architectures}

We experimented with two models: a baseline 1D ResNet-like model \citep{HeEtAl2016} and the state-of-the-art Fourier Neural Operator (FNO) network \citep{li2020fourier}.
Both models take $K$ history timesteps as input and are trained in an autoregressive way and as neural operator (see Figure \ref{fig:schematic}).
The input therefore has the dimension $[\text{batch}, K, c]$ where $K$ is interpreted as channel dimension. The channel dimension is mapped to $128$ and $256$ dimension for the 1D ResNet and the FNO, respectively.
The 1D ResNet has 8 convolutional layers using residual connections and increasing receptive fields (kernel size 3 to kernel size 15) and ELU \citep{clevert2015fast} non-linearities in between. The FNO architectures use 5 1D FNO layers as proposed in \citet{li2020fourier} with 32 Fourier modes. These layers have residual connections and are intertwined with GeLU \citep{hendrycks2016gaussian} non-linearieties. We also test a 2D FNO architecture as alternatively described in 
\citet{li2020fourier}, but the 1D FNO worked consistently better.
For both models, the output is mapped to $K$ channels in the autoregressive setting and to the number of desired timesteps in the neural operator setting.

\paragraph{A note on Fourier Neural Operators.}
The way we interpret the Fourier Neural Operator (FNO) is via its different treatment of the spatial and temporal informations.
FNO layers~\citep{li2020fourier} consist of two parts, which are convolutions in the Fourier domain and convolutions with kernel size equals one. 
Whereas the former focus on processing spatial information, the latter process temporal information if temporal information is provided in the channel dimension. We therefore merge the temporal bundling idea of~\citet{BrandstetterEtAl2022} into the FNO framework, where $K$ timesteps are input to the model. Consequently, the temporal information processing is strongly enhanced. This is not only results in improved runtimes by a factor $K$ for rollouts, but also in better stability and decreased errors, see e.g. Figure~\ref{fig:long_rollouts} and Tables~\ref{tab:long_rollouts_KdV} and \ref{tab:long_rollouts_KS}.

\paragraph{Training details.}
We optimize models using the AdamW optimizer~\citep{loshchilov2017decoupled} with learning rate 1e-4, weight decay 1e-8 for 20 epochs and minimize the normalized mean squared error (NMSE) which is outlined in Equation \ref{eq:loss}.
The overall used FNO architectures consist of roughly 1 million parameters and training for the different experiments takes between 12 and 24 hours on average on a GeForceRTX 2080 Ti GPU.

\section{Results} \label{sec:app:results}

Table \ref{tab:summary_of_results} and Table \ref{tab:summary_of_results_Burgers} contain a comprehensive summary of results presented in Figures \ref{fig:kdv-ks}, \ref{fig:kdv-model}, \ref{fig:kdv-method}, \ref{fig:kdv-multiple}, i.e. an extensive test of Lie point symmetry data augmentation (LPSDA) on the Korteweg-de Vries (KdV) and the Kuramoto-Shivashinsky (KS) equation.
Table \ref{tab:long_rollouts_KdV} and Table \ref{tab:long_rollouts_KS} contain a comprehensive summary of KdV and KS long rollout results, discussed in Section \ref{sec:very-long} in the main paper.

\onecolumn

\begin{table*}[!htb]
\caption{Test of Lie point symmetry data augmentation (LPSDA) on the Korteweg-de Vries (KdV) and the Kuramoto-Shivashinsky (KS) equation. Averaged normalized MSE (NMSE) errors on the test set are reported, $\gL_{\text{NMSE}} \frac{1}{N_t}\sum_{j=1}^{N_t} \frac{\Vert \rvu(t_j) - \hat{\rvu}(t_j) \Vert_2^2}{\Vert \hat{\rvu}(t_j) \Vert^2}$.
Autoregressive (AR) and neural operator (NO) prediction errors 
are compared. A 1D ResNet-like and a Fourier Neural Operator \citet{li2020fourier} (FNO) solver are compared. Best performing symmetry groups are highlighted.
Different solvers are tested on different combination of Lie point symmetries. Experiments show that performance gets consistently better for additional generators. This is verified for different solvers, different equations, different tasks, and different combination of generators.
}
\label{tab:summary_of_results}
\setlength\tabcolsep{0pt} 
\scriptsize\centering
\smallskip 
\resizebox{\columnwidth}{!}{
\begin{tabular*}{\columnwidth}{@{\extracolsep{\fill}}ccc | cc cc cc}
\toprule
& & & \multicolumn{6}{c}{Number of samples}\\
\midrule
Task & Solver & Sym. & 512 & 256 & 128 & 64 & 32 & 16 \\
\midrule
KdV ($20s$) & FNO (AR) & - & $.0030\pm.0010$ & $.0058\pm.0010$ & $.0225\pm.0024$ & $.0604\pm.0036$ & $.1328\pm.0093$ & $.1901\pm.0105$\\
KdV ($20s$) & FNO (AR) & $g_1 g_2 g_3 g_4$ & $\bm{.0010\pm.0001}$ & $\bm{.0037\pm.0006}$ & $\bm{.0119\pm.0013}$ & $\bm{.0387\pm.0025}$ & $\bm{.0959\pm.0082}$ & $\bm{.1399\pm.0065}$ \\
\midrule
KdV ($20s$) & FNO (NO) & - & $.0276\pm.0038$ & $.0407\pm.0037$ & $.0858\pm.0056$ & $.1699\pm.0100$ & $.2717\pm.0146$ & $.4887\pm.0156$ \\
KdV ($20s$) & FNO (NO) & $g_1 g_2 g_3 g_4$ & $\bm{.0055\pm.0007}$ & $\bm{.0132\pm.0013}$ & $\bm{.0302\pm.0021}$ & $\bm{.0574\pm.0045}$ & $\bm{.1289\pm.0084}$ & $\bm{.2376\pm.0136}$ \\
\midrule
\midrule
KdV ($20s$) & ResNet (NO) & - & $.0160\pm.0014$ & $.0273\pm.0014$ & $.0443\pm.0026$ & $.0574\pm.0030$ & $.1021\pm.0056$ & $.1178\pm.0064$\\
KdV ($20s$) & ResNet (NO) & $g_1 g_2$ & $.0161\pm.0014$ & $.0231\pm.0023$ & $.0291\pm.0031$ & $.0434\pm.0025$ & $.0896\pm.0072$ & $.1239\pm.0095$ \\
KdV ($20s$) & ResNet (NO) & $g_1g_2 g_3$ & $.0168\pm.0014$ & $\bm{.0209\pm.0011}$ & $.0310\pm.0024$ & $.0426\pm.0021$ & $.0824\pm.0049$ & $.0990\pm.0037$\\
KdV ($20s$) & ResNet (NO) & $g_1 g_2 g_4$ & $\bm{.0121\pm.0013}$ & $\bm{.0204\pm.0016}$ & $\bm{.0276\pm.0021}$ & $.0474\pm.0021$ & $.1314\pm.0077$ & $.1188\pm.0054$ \\
KdV ($20s$) & ResNet (NO) & $g_1 g_2 g_3 g_4$ & $\bm{.0118\pm.0010}$ & $\bm{.0205\pm.0016}$ & $\bm{.0277\pm.0020}$ & $\bm{.0407\pm.0026}$ & $\bm{.0774\pm.0048}$ & $\bm{.0746\pm.0036}$ \\
\midrule
KdV ($20s$) & ResNet (AR) & - & $.0223\pm.0026$ & $.0392\pm.0031$ & $.0669\pm.0050$ & $.0914\pm.0087$ & $.1701\pm.0110$ & $.2087\pm.0121$\\
KdV ($20s$) & ResNet (AR) & $g_1 g_2$ & $.0200\pm.0021$ & $.0284\pm.0022$ & $.0470\pm.0031$ & $.0782\pm.0049$ & $.1463\pm.0099$ & $.1911\pm.0110$ \\
KdV ($20s$) & ResNet (AR) & $g_1g_2 g_3$ & $.0111\pm.0010$ & $.0185\pm.0013$ & $.0318\pm.0022$ & $.0517\pm.0029$ & $.0883\pm.0068$ & $.1145\pm.0069$\\
KdV ($20s$) & ResNet (AR) & $g_1 g_2 g_4$ & $.0155\pm.0019$ & $.0269\pm.0026$ & $.0445\pm.0028$ & $.0763\pm.0045$ & $.1602\pm.0104$ & $.1887\pm.0088$ \\
KdV ($20s$) & ResNet (AR) & $g_1 g_2 g_3 g_4$ & $\bm{.0113\pm.0012}$ & $\bm{.0184\pm.0016}$ & $\bm{.0333\pm.0020}$ & $\bm{.0544\pm.0030}$ & $\bm{.0965\pm.0064}$ & $\bm{.1106\pm.0052}$ \\
\midrule
\midrule
KdV ($40s$) & FNO (AR) & - & $.0081\pm.0026$ & $.0361\pm.0062$ & $.0681\pm.0084$ & $.1248\pm.0108$ & $.1981\pm.0167$ & $.2904\pm.0184$\\
KdV ($40s$) & FNO (AR) & $g_1$ & $.0023\pm.0004$ & $.0082\pm.0010$ & $.0293\pm.0024$ & $ .0674\pm.0055$ & $.1294\pm.0081$ & $.2224\pm.0123$ \\
KdV ($40s$) & FNO (AR) & $g_1 g_2$ & $.0016\pm.0008$ & $.0088\pm.0017$ & $.0278\pm.0045$ & $ .0673\pm.0058$ & $.1251\pm.0106$ & $.2160\pm.0135$ \\
KdV ($40s$) & FNO (AR) & $g_1g_2 g_3$ & $\bm{.0011\pm.0005}$ & $.0062\pm.0012$ & $\bm{.0229\pm.0035}$ & $\bm{.0550\pm.0058}$ & $\bm{.0940\pm.0079}$ & $\bm{.1818\pm.0153}$\\
KdV ($40s$) & FNO (AR) & $g_1 g_2 g_4$ & $.0017\pm.0009$ & $.0102\pm.0016$ & $.0287\pm.0048$ & $.0795\pm.0077$ & $.1271\pm.0096$ & $.2092\pm.0138$ \\
KdV ($40s$) & FNO (AR) & $g_1 g_2 g_3 g_4$ & $\bm{.0010\pm.0006}$ & $\bm{.0058\pm.0013}$ & $\bm{.0213\pm.0029}$ & $.0606\pm.0040$ & $\bm{.0968\pm.0091}$ & $\bm{.1817\pm.0118}$ \\
\midrule
KdV ($40s$) & FNO (NO) & - & $.1113\pm.0100$ & $.1761\pm.0122$ & $.2343\pm.0194$ & $.3331\pm.0185$ & $.4892\pm.0237$ & $.5954\pm.0301$\\
KdV ($40s$) & FNO (NO) & $g_1$ & $.0735\pm.0040$ & $.1162\pm.0053$ & $.1723\pm.0083$ & $.2350\pm.0076$ & $.3674\pm.0149$ & $.4627\pm.0112$ \\
KdV ($40s$) & FNO (NO) & $g_1 g_2$ & $.0761\pm.0086$ & $.1140\pm.0087$ & $.1622\pm.0121$ & $.2494\pm.0158$ & $.3432\pm.0232$ & $.4884\pm.0254$ \\
KdV ($40s$) & FNO (NO) & $g_1g_2 g_3$ & $\bm{.0279\pm.0038}$ & $\bm{.0462\pm.0052}$ & $\bm{.0868\pm.0072}$ & $\bm{.1170\pm.0111}$ & $\bm{.1842\pm.0130}$ & $\bm{.3001\pm.0169}$\\
KdV ($40s$) & FNO (NO) & $g_1 g_2 g_4$ & $.0779\pm.0086$ & $.1140\pm.0098$ & $.1757\pm.0130$ & $.2534\pm.0139$ & $.3731\pm.0207$ & $.4764\pm.0204$ \\
KdV ($40s$) & FNO (NO) & $g_1 g_2 g_3 g_4$ & $\bm{.0273\pm.0040}$ & $\bm{.0478\pm.0052}$ & $\bm{.0873\pm.0070}$ & $\bm{.1127\pm.0101}$ & $\bm{.1743\pm.0118}$ & $\bm{.3037\pm.0190}$ \\
\midrule
\midrule
KS ($20s$) & FNO (AR) & - & $.0279\pm.0013$ & $.1084\pm.0048$ & $.1983\pm.0073$ & $.2898\pm.0080$ & $.4234\pm.0106$ & $.5279\pm.0110$\\
KS ($20s$) & FNO (AR) & $g_1$ & $.0045\pm.0004$ & $.0421\pm.0022$ & $.1347\pm.0045$ & $.2241\pm.0067$ & $.3690\pm.0086$ & $.4984\pm.0143$\\
KS ($20s$) & FNO (AR) & $g_1g_2$ & $.0057\pm.0004$ & $.0581\pm.0020$ & $.1321\pm.0056$ & $.2316\pm.0085$ & $.3605\pm.0109$ & $.4945\pm.0122$ \\
KS ($20s$) & FNO (AR) & $g_1 g_2 g_3$ & $\bm{.0028\pm.0002}$ & $\bm{.0261\pm.0012}$ & $\bm{.0755\pm.0035}$ & $\bm{.1504\pm.0064}$ & $\bm{.2656\pm.0074}$ & $\bm{.4398\pm.0133}$ \\
\midrule
KS ($20s$) & FNO (NO) & - & $.3905\pm.0049$ & $.4733\pm.0051$ & $.5662\pm.0051$ & $.6855\pm.0097$ & $.8502\pm.0064$ & $1.0005\pm.0076$\\
KS ($20s$) & FNO (NO) & $g_1$ & $.2936\pm.0044$ & $.3863\pm.0050$ & $.4813\pm.0050$ & $.5884\pm.0057$ & $.7438\pm.0054$ & $.8426\pm.0060$ \\
KS ($20s$) & FNO (NO) & $g_1g_2$ & $.2948\pm.0047$ & $.3864\pm.0043$ & $.4854\pm.0043$ & $.5810\pm.0061$ & $.7388\pm.0048$ & $.8359\pm.0070$\\
KS ($20s$) & FNO (NO) & $g_1 g_2 g_3$ & $\bm{.2268\pm.0049}$ & $\bm{.3089\pm.0066}$ & $\bm{.3981\pm.0075}$ & $\bm{.4501\pm.0069}$ & $\bm{.5321\pm.0038}$ & $\bm{.6045\pm.0057}$ \\
\bottomrule
\end{tabular*}
}
\end{table*}

\begin{table*}[!htb]
\caption{Test of Lie point symmetry data augmentation (LPSDA) on the Burgers' equation. Averaged normalized MSE (NMSE) errors on the test set are reported, $\gL_{\text{NMSE}} \frac{\Vert \rvu(t_j) - \hat{\rvu}(t_j) \Vert_2^2}{\Vert \hat{\rvu}(t_j) \Vert^2}$ is omitted due to the small values. Autoregressive (AR) and neural operator (NO) prediction errors 
are compared. Fourier Neural Operator \citet{li2020fourier} (FNO) solver are tested on different combination of Lie point symmetries. Best performing symmetry groups are highlighted. Experiments show that performance gets consistently better for additional generators. 
}

\label{tab:summary_of_results_Burgers}
\setlength\tabcolsep{0pt} 
\scriptsize\centering

\smallskip 
\begin{tabular*}{\columnwidth}{@{\extracolsep{\fill}}ccc | cc cc cc}
\toprule
& & & \multicolumn{6}{c}{Number of samples}\\
\midrule
Task & Solver & Sym. & 512 & 256 & 128 & 64 & 32 & 16 \\
\midrule
Burgers' ($10s$) & FNO (AR) & - & $.0020\pm.0002$ & $.0091\pm.0008$ & $.0696\pm.0057$ & $.3442\pm.0226$ & $.7962\pm.0358$ & $1.6493\pm.0581$ \\
Burgers' ($10s$) & FNO (AR) & $g_1 $ & $.0010\pm.0001$ & $.0061\pm.0006$ & $.0455\pm.0034$ & $.2767\pm.0189$ & $.8489\pm.0461$ & $2.5599\pm.0954$\\
Burgers' ($10s$) & FNO (AR) & $g_1g_5$ & $\mathbf{.0003\pm.0001}$ & $\bm{.0010\pm.0002}$ & $\bm{.0063\pm.0014}$ & $\bm{.0228\pm.0031}$ & $\bm{.0656\pm.0104}$ & $\bm{.3729\pm.0303}$ \\
\midrule
Burgers' ($10s$) & FNO (NO) & - & $.0428\pm.0049$ & $.1145\pm.0110$ & $.2792\pm.0199$ & $.5692\pm.0306$ & $.7959\pm.0409$ & $1.8415\pm.0645$\\
Burgers' ($10s$) & FNO (NO) & $g_1 $ & $.0092\pm.0014$ & $.0566\pm.0050$ & $.2723\pm.0156$ & $.6092\pm.0290$ & $.9784\pm.0445$ & $2.3453\pm.0663$\\
Burgers' ($10s$) & FNO (NO) & $g_1g_5$ & $\bm{.0003\pm.0001}$ & $\bm{.0009\pm.0001}$ & $\bm{.0066\pm.0015}$ & $\bm{.0322\pm.0043}$ & $\bm{.1436\pm.0112}$ & $\bm{.5511\pm.0281}$ \\
\bottomrule
\end{tabular*}
\end{table*}

\begin{table*}[!htb]
\caption{Long rollouts of $T=200\text{ s}$ with $N_{t_{out}}=400$ timesteps on the Korteweg-de Vries (KdV) equation to check how LPSDA can improve performance in a practical use case. The cumulative error normalized over seconds is reported. We use an autoregressively-trained FNO model. Additionally, the training procedure of the solvers is augmented with temporal bundling (TB) and the pushforward trick (PF) as introduced in \citet{BrandstetterEtAl2022}. For temporal bundling variants we input and output 20 timesteps at a time, which we compare to one timestep predictions (1-step). All training tricks are ablated. Best performing symmetry groups are highlighted.}

\label{tab:long_rollouts_KdV}
\setlength\tabcolsep{0pt} 
\scriptsize\centering

\smallskip 
\begin{tabular*}{0.8\columnwidth}{@{\extracolsep{\fill}}c|cccc}
\toprule
& \multicolumn{4}{c}{Training methods}\\
\midrule
Timesteps & 1-step & TB & TB+PF & TB+PF+LPSDA \\
\midrule
20s & $.0008\pm.0005$ & $.0008\pm.0005$ & $.0011\pm.0005$ & $\mathbf{.0003\pm.0001}$\\
40s & $.0046\pm.0011$ & $.0065\pm.0021$ & $.0048\pm.0016$ & $\mathbf{.0012\pm.0003}$\\
60s & $.0120\pm.0021$ & $.0176\pm.0046$ & $.0119\pm.0034$ & $\mathbf{.0028\pm.0009}$\\
80s & $.0242\pm.0037$ & $.0287\pm.0065$ & $.0209\pm.0052$ & $\mathbf{.0053\pm.0013}$\\
100s & $.0441\pm.0084$ & $.0423\pm.0083$ & $.0326\pm.0064$ & $\mathbf{.0094\pm.0018}$\\
120s & $.0628\pm.0032$ & $.0610\pm.0110$ & $.0446\pm.0080$ & $\mathbf{.0151\pm.0024}$\\
140s & $.0946\pm.0074$ & $.0744\pm.0125$ & $.0616\pm.0102$ & $\mathbf{.0231\pm.0040}$\\
160s & $.1345\pm.0093$& $.0932\pm.0150$ & $.0801\pm.0123$ & $\mathbf{.0338\pm.0047}$\\
180s & $\infty$ & $.1167\pm.0181$ & $.0959\pm.0137$ & $\mathbf{.0454\pm.0048}$\\
200s & $\infty$ & $.1369\pm.0188$ & $.1198\pm.0136$ & $\mathbf{.0621\pm.0056}$\\
\bottomrule
\end{tabular*}
\end{table*}

\begin{table*}[!htb]
\caption{Long rollouts of $T=100\text{ s}$ with $N_{t_{out}}=400$ timesteps on the Kuramoto-Shivashinsky (KS) equation to check how LPSDA can improve performance in a practical use case. The cumulative error normalized over seconds is reported. We use an autoregressively-trained FNO model. Additionally, the training procedure of the solvers is augmented with temporal bundling (TB) and the pushforward trick (PF) as introduced in \citet{BrandstetterEtAl2022}. For temporal bundling variants we input and output 20 timesteps at a time, which we compare to one timestep predictions (1-step). All training tricks are ablated. Best performing symmetry groups are highlighted.}

\label{tab:long_rollouts_KS}
\setlength\tabcolsep{0pt} 
\scriptsize\centering

\smallskip 
\begin{tabular*}{0.8\columnwidth}{@{\extracolsep{\fill}}c|rrrrr}
\toprule
& \multicolumn{5}{c}{Training methods}\\
\midrule
Timesteps & 1-step & TB & TB+PF & TB+PF+LPSDA & TB+LPSDA\\
\midrule
10s & $.0003\pm.0001$ & $.0001\pm.0000$ & $.0001\pm.0000$ & $\mathbf{.0000\pm.0000}$ & $\mathbf{.0000\pm.0000}$ \\
20s & $.0060\pm.0010$ & $.0009\pm.0001$ & $.0011\pm.0001$ & $.0003\pm.0000$ & $\mathbf{.0002\pm.0000}$ \\
30s & $.0283\pm.0040$ & $.0077\pm.0013$ & $.0080\pm.0013$ & $.0024\pm.0002$ & $\mathbf{.0017\pm.0002}$ \\
40s & $.0391\pm.0359$ & $.0377\pm.0054$ & $.0381\pm.0060$ & $.0126\pm.0021$ & $\mathbf{.0101\pm.0019}$ \\\
50s & $\infty$ & $.1166\pm.0128$ & $.1126\pm.0138$ & $.0536\pm.0051$ & $\mathbf{.0419\pm.0049}$\\
60s & $\infty$ & $.2914\pm0222$ & $.2793\pm.0224$ & $.1415\pm.0127$ & $\mathbf{.1278\pm.0110}$\\
70s & $\infty$ & $.5604\pm0273$ & $.5459\pm.0283$ & $.3088\pm.0188$ & $\mathbf{.2853\pm.0190}$\\
80s & $\infty$ & $.9387\pm.4007$ & $.9011\pm.0393$ & $.5668\pm.0256$ & $\mathbf{.5241\pm.0264}$\\
90s & $\infty$ & $1.3422\pm.0523$ & $1.3049\pm.0598$ & $.9369\pm.0392$ & $\mathbf{.8736\pm.0390}$\\
100s & $\infty$ & $1.7821\pm.0564$ & $1.7431\pm.0587$ & $1.3692\pm.0572$ & $\mathbf{1.2741\pm.0570}$ \\
\bottomrule
\end{tabular*}
\end{table*}


\end{document}